\documentclass[hidelinks]{article}

\newif\ifanonymous

\usepackage{arxiv}

\usepackage{blindtext}
\usepackage{graphicx}
\usepackage{hyperref}
\usepackage{url}
\usepackage{optidef}
\usepackage{times}
\usepackage{caption}
\usepackage{subcaption}
\usepackage[super]{nth}
\usepackage{ifdraft}
\usepackage{multirow}
\usepackage{placeins}
\usepackage{comment}
\usepackage{tabularx}
\usepackage[colorinlistoftodos,prependcaption]{todonotes}
\usepackage{amssymb}
\usepackage{amsthm}
\usepackage{algorithm}
\usepackage{algpseudocode}

\graphicspath{{./figures/}}

\usepackage{amsmath,amsfonts,bm,mathtools,xargs}

\def\Tableref#1{Table~\ref{#1}}
\def\Twotablerefs#1#2{Table~\ref{#1} and~\ref{#2}}

\def\Figref#1{Figure~\ref{#1}}

\def\secref#1{section~\ref{#1}}
\def\Secref#1{Section~\ref{#1}}

\def\Twosecrefs#1#2{Sections~\ref{#1} and~\ref{#2}}

\def\eqref#1{equation~\ref{#1}}

\def\Eqref#1{Equation~\ref{#1}}

\def\Algref#1{Algorithm~\ref{#1}}

\def\Appref#1{Appendix~\ref{#1}}

\def\Twoappref#1#2{Appendices~\ref{#1} and~\ref{#2}}
\def\Thref#1{Theorem~\ref{#1}}

\def\Axref#1{Axiom~\ref{#1}}

\def\Remref#1{Remark~\ref{#1}}

\def\1{\bm{1}}

\def\rvx{{\mathbf{x}}}

\def\rvz{{\mathbf{z}}}

\def\vmu{{\bm{\mu}}}
\def\vtheta{{\bm{\theta}}}
\def\vphi{{\bm{\phi}}}

\def\evsigma{{\sigma}}

\def\mI{{\bm{I}}}

\def\mX{{\bm{X}}}
\def\mY{{\bm{Y}}}

\DeclareMathAlphabet{\mathsfit}{\encodingdefault}{\sfdefault}{m}{sl}
\SetMathAlphabet{\mathsfit}{bold}{\encodingdefault}{\sfdefault}{bx}{n}

\def\sA{{\mathbb{A}}}
\def\sB{{\mathbb{B}}}

\def\sS{{\mathbb{S}}}

\newcommand{\N}{\mathcal{N}}

\newcommand{\E}{\mathbb{E}}

\newcommand{\R}{\mathbb{R}}

\newcommand{\KL}{D_{\mathrm{KL}}}
\newcommandx{\D}[5][1,2]{D^{#1}_{#2{#3}}(#4||#5)}

\DeclareMathOperator{\ELBO}{\mathcal{L}(\vtheta, \vphi; \rvx)}

\newtheorem{axm}{Axiom}
\newtheorem{thm}{Theorem}
\newtheorem{rem}{Remark}

\newcommandx{\suggestion}[2][1=]{\todo[linecolor=ProcessBlue,backgroundcolor=ProcessBlue!25,bordercolor=ProcessBlue,#1]{#2}}
\newcommandx{\donelast}[2][1=]{\todo[linecolor=gray,backgroundcolor=gray!25,bordercolor=gray,#1]{#2}}
\renewcommand\labelenumi{(\roman{enumi})}
\renewcommand\theenumi\labelenumi

\anonymousfalse

\title{Fondue: an algorithm to find the optimal dimensionality of the latent representations of variational autoencoders}

\author{Lisa Bonheme \& Marek Grzes
    School of Computing\\
    University of Kent\\
    Canterbury, UK\\
    \texttt{\{lb732, m.grzes\}@kent.ac.uk}
}

\begin{document}
    \maketitle

    \begin{abstract}
        When training a variational autoencoder (VAE) on a given dataset, determining the optimal number of latent
        variables is mostly done by grid search --- a costly process in terms of computational time and carbon footprint.
        In this paper, we explore the intrinsic dimension estimation (IDE) of the data and latent representations learned by VAEs.
        We show that the discrepancies between the IDE of the mean and sampled representations of a VAE after only a few steps of
        training reveal the presence of passive variables in the latent space, which,
        in well-behaved VAEs, indicates a superfluous number of dimensions.
        Using this property, we propose FONDUE: an algorithm which quickly finds the number of latent dimensions
        after which the mean and sampled representations start to diverge (i.e., when passive variables are introduced), providing a principled method for selecting the
        number of latent dimensions for VAEs and autoencoders.
    \end{abstract}

    \section{Introduction}\label{sec:intro}
``How many latent variables should I use for this model?'' is a question that many practitioners using variational autoencoders (VAEs) or autoencoders (AEs) have to deal with.
When the task has been studied before, this information is available in the literature for the specific architecture and dataset used.
However, when it has not, answering this question becomes more complicated.
Indeed, the dimensionality of the latent representation is usually determined empirically
by increasing the number of latent dimensions until the reconstruction does not improve anymore~\citep{Doersch2016}.
This is a costly process requiring multiple model training, and increasing the carbon footprint and time needed for an experiment.

In recent years, topology-based methods have successfully been applied to deep learning~\citep{Hensel2021} and generative models to design new
metrics~\citep{Khrulkov2018,Zhou2021,Rieck2018} and learning methods~\citep{Falorsi2018,PerezRey2020,Keller2021}.
They have also been used to analyse the representations learned by deep neural networks (DNNs)~\citep{Arvanitidis2018,Ansuini2019,Maheswaranathan2019,Naitzat2020}.

Intrinsic dimension estimation (IDE) --- the estimation of the minimum number of variables needed to
describe the data --- is an active area of reseach in topology, and various estimation methods have been proposed~\citep{Facco2017,Levina2004}.
Using these techniques, the intrinsic dimension (ID) of images was empirically shown to be
much lower than their extrinsic dimension (i.e., the number of pixels)~\citep{Gong2019,Ansuini2019,Pope2021}.
Moreover,~\citet{Ansuini2019} observed that the ID of neural network classifiers with good generalisation tended to first increase,
then decrease until reaching a very low ID in the last layer.

As IDE provides an estimate of the minimum number of variables needed to describe the data, it could
be an invaluable tool to determine the number of latent variables needed for VAEs and avoid costly grid searches,
effectively reducing the carbon footprint of model implementations.

Thus, our objective in this paper is to verify whether the ID of different layers can be used to determine the number of latent variables in VAEs.

Our contributions are as follows:
\begin{enumerate}
    \itemsep 0em
    \item We provide an experimental study of the IDE of VAEs, and have released more than 35,000 IDE scores
          \ifanonymous\else(\url{https://data.kent.ac.uk/id/eprint/455}\fi).
    \item We have released the library created for this experiment (\ifanonymous\else\url{https://github.com/bonheml/VAE_learning_dynamics}\fi).
          It can be reused with other IDE techniques or models for further research in the domain.
    \item During our analysis of VAEs, we found that (1) the deeper the layer of the encoder, the lower the estimated IDs,
          whereas the layers of the decoder all have the same IDE;
          (2) the extrinsic dimensionality of the latent representations is generally higher than its IDE; (3) the layers reach
          a stable ID very early in the training; and (4) the IDE of mean and sampled representations is different when some latent variables collapse.
    \item Based on these findings, we propose FONDUE: an algorithm for Finding the Optimal Number of Dimensions by Unsupervised Estimation, which works well
          on the three datasets used in our experiment.
\end{enumerate}

    \section{Background}\label{sec:background}

\subsection{Variational Autoencoders}\label{subsec:bg-VAEs}
VAEs~\citep{Kingma2013,Rezende2015} are deep probabilistic generative models based
on variational inference.~The encoder maps an input $\rvx$ to a latent representation $\rvz$, and
the decoder attempts to reconstruct $\rvx$ using  $\rvz$.
This can be optimised by maximising $\mathcal{L}$, the evidence lower bound (ELBO)
\begin{equation}
    \label{eq:elbo}
    \ELBO = \underbrace{\E_{q_\vphi(\rvz|\rvx)}[\log p_\vtheta(\rvx|\rvz)]}_{\text{reconstruction term}} -
    \underbrace{\KL\left(q_\vphi(\rvz|\rvx) || p(\rvz)\right)}_{\text{regularisation term}},
\end{equation}
where $p(\rvz)$ is generally modelled as a standard multivariate Gaussian distribution $\N(0, \mI)$ to permit a closed
form computation of the regularisation term~\citep{Doersch2016}.
The regularisation term can be further penalised by a weight $\beta$~\citep{Higgins2017} such that
\begin{equation}
    \label{eq:beta-vae}
    \ELBO = \underbrace{\E_{q_\vphi(\rvz|\rvx)}[\log p_\vtheta(\rvx|\rvz)]}_{\text{reconstruction term}} -
    \underbrace{\beta\KL\left(q_\vphi(\rvz|\rvx) || p(\rvz)\right)}_{\text{regularisation term}},
\end{equation}
reducing to~\eqref{eq:elbo} when $\beta=1$ and to a deterministic auto-encoder when $\beta=0$.

\paragraph{Posterior collapse and polarised regime}
When $\beta \geqslant 1$, VAEs can produce disentangled representations~\citep{Higgins2017} but too high values of $\beta$ result in
posterior collapse (i.e., $\rvz \sim \N(0, \mI)$) making the model unusable~\citep{Lucas2019,Lucas2019b,Dai2020}.
Indeed, the sampled representation will not retain any information from the input, making it impossible for the decoder to correctly reconstruct the image.
Nevertheless, for VAEs to provide good reconstruction~\citep{Dai2018,Dai2020}, it is necessary for any superfluous dimensions of
$\rvz$ to be collapsed and ignored by the decoder. These collapsed dimensions are called passive variables; the remaining,
active variables. When this selective posterior collapse behaviour
--- also known as the polarised regime~\citep{Rolinek2019} --- happens, the passive variables are very
different in mean and sampled representations~\citep{Bonheme2021} (see also \Twoappref{sec:app-vae}{sec:app-collapse}).

\subsection{Intrinsic dimension estimation}\label{subsec:bg-ide}
It is generally assumed that a dataset $\mX$ of $m$ i.i.d. data examples $\mX_i \in \R^n$ is a locally smooth non-linear
transformation $g$ of a lower-dimensional dataset $\mY$ of $m$ i.i.d. samples $\mY_i \in \R^d$,
where $d \leqslant n$~\citep{Campadelli2015, Chollet2021}.~The goal of intrinsic dimension estimation (IDE) is to recover $d$
given $\mX$.~In this section, we will detail two IDE techniques which use the statistical properties of the
neighbourhood of each data point to estimate $d$, and provide good results for approximating the ID of deep neural
network representations and deep learning datasets~\citep{Ansuini2019,Gong2019,Pope2021}.

\paragraph{Maximum Likelihood Estimation}
\citet{Levina2004} modelled the neighbourhood of a given point $\mX_i$ as a Poisson process in a d-dimensional sphere
$S_{\mX_i}(R)$ of radius $R$ around $\mX_i$. This Poisson process is denoted $\{N(t, \mX_i), 0 \leqslant t \leqslant R\}$, where
$N(t, \mX_i)$ is a random variable (distributed according to a Poisson distribution) representing the number of
neighbours of $\mX_i$ within a radius $t$. Each point $\mX_j \in S_{\mX_i}(R)$ is thus considered as an event, its arrival
time $t=T(\mX_i, \mX_j)$ being the Euclidean distance from $\mX_{i}$ to its $j^{th}$ neighbour $\mX_j$.
By expressing the rate $\lambda(t, \mX_i)$ of the process $N(t, \mX_i)$ as a
function of the surface area of the sphere --- and thus relating $\lambda(t, \mX_i)$ to $d$ ---
they obtain a maximum likelihood estimation (MLE) of the intrinsic dimension $d$:
\begin{equation}
    \label{eq:ide-mle-poisson}
    \bar{d}_{R}(\mX_i) = \left[ \frac{1}{N(R, \mX_i)}\sum_{j=1}^{N(R, \mX_i)} \log \frac{R}{T(\mX_i, \mX_j)} \right]^{-1}.
\end{equation}
\Eqref{eq:ide-mle-poisson} is then simplified by fixing the number of neighbours, $k$, instead of the radius $R$ of the
sphere, such that
\begin{equation}
    \label{eq:ide-mle}
    \bar{d}_{k}(\mX_i) = \left[ \frac{1}{k-1}\sum_{j=1}^{k-1} \log \frac{T(\mX_i, \mX_k)}{T(\mX_i, \mX_j)} \right]^{-1},
\end{equation}
where the last summand is omitted, as it is zero for $j=k$.
The final estimate $\bar{d}_k$ is the averaged score over $n$ data examples~\citep{Levina2004}
\begin{equation}
    \label{eq:ide-mle-avg}
    \bar{d}_k = \frac{1}{n} \sum_{i=1}^n \bar{d}_{k}(\mX_i).
\end{equation}

To obtain an accurate estimation of the ID with MLE, it is very important to choose a sufficient number of neighbours
$k$ to form a dense small sphere~\citep{Levina2004}.~On one hand, if $k$ is too small, MLE will generally
underestimate the ID, and suffer from high variance~\citep{Levina2004,Campadelli2015,Pope2021}.
On the other hand, if $k$ is too large, the ID will be overestimated~\citep{Levina2004,Pope2021}.

\paragraph{TwoNN}
\citet{Facco2017} proposed an estimation of the ID based on the ratio of the two nearest neighbours of
$\mX_i$, $r_{\mX_i} = \frac{T(\mX_i, \mX_l)}{T(\mX_i, \mX_j)}$, where $\mX_j$ and $\mX_l$ are the first and second closest
neighbours of $\mX_i$, respectively.
$r$ follows a Pareto distribution with scale $s = 1$ and shape $d$, and its density function $f(r)$ is
\begin{equation}
    \label{eq:twonn-pdf}
    f(r) = \frac{d s^{d}}{r^{d+1}} = d r^{-(d+1)}.
\end{equation}
Its cumulative distribution function is thus
\begin{equation}
      \label{eq:twonn-cdf}
      F(r) = 1 - \frac{s^{d}}{r^{d}} = 1 - r^{-d},
\end{equation}
and, using \Eqref{eq:twonn-cdf}, one can readily obtain $d = \frac{- \log(1 - F(r))}{\log{r}}$.
From this, we can see that $d$ is the slope of the straight line passing through the origin, which is given by the set of coordinates
$\sS = \{ ( \: \log r_{\mX_i}, - \log (1 - F(r_{\mX_i})) \: ) \;|\; i = 1, \cdots, m\}$,
and can be recovered by linear regression.

As TwoNN uses only two neighbours, it can be sensitive to outliers~\citep{Facco2017} and do not perform well on
high ID~\citep{Pope2021}, overestimating the ID in both cases.

\paragraph{Ensuring an accurate analysis}~Given the limitations previously mentioned, we take two remedial actions
to guarantee that our analysis is as accurate as possible.~To provide an IDE which is as accurate as possible with MLE,
we will measure the MLE with an increasing number of neighbours and, similar to~\citet{Karbauskaita2011}, retain
the IDE which is stable for the largest number of $k$ values.~TwoNN will be used as a complementary metric to validate
our choice of $k$ for MLE. In case of significant discrepancies with a significantly higher TwoNN IDE, we will rely on
the results provided by MLE.

    \section{Experimental setup}\label{sec:xp}
As mentioned in \secref{sec:intro}, the main objective of our experiment is to investigate the IDs of the
representations learned by VAEs to assess whether they can be used to determine the optimal number of latent dimensions of VAEs.

To do so, we will train VAEs with at least 8 different numbers of latent dimensions on 3 datasets of increasing ID and estimate
the ID of each layer of the models.~We will then analyse these results and use
them to verify our objectives in~\Secref{sec:res}.

\paragraph{Datasets} We use three datasets with an increasing number of intrinsic dimensions:
Symmetric solids~\citep{Murphy2021}, dSprites~\citep{Higgins2017}, and Celeba~\citep{Liu2015}.
The numbers of generative factors of the first two datasets are 2 and 5, respectively,
and the IDE of these two datasets should be close to these values.
While we do not know the generative factors of Celeba,~\citet{Pope2021} reported an
IDE greater than 20, which is high enough for our experiment.

\paragraph{Data preprocessing} Each image is resized to $64 \times 64 \times c$, where $c=1$ for Symmetric solids and
dSprites, and $c=3$ for Celeba. We also removed duplicate images (i.e., cases where different rotations resulted in the
same image) and labels from Symmetric solids and created a reduced version: \texttt{symsol\_reduced} which is available
\ifanonymous\else~at \url{https://data.kent.ac.uk/436}\fi.

\paragraph{VAE training}
We use the $\beta$-VAE architecture detailed in~\citet{Higgins2017} for all the datasets, together with the standard learning
objective of VAEs, as presented in~\Eqref{eq:elbo}.
Each VAE is trained 5 times with a number of latent dimensions $n=3,6,8,10,12,18,24,32$ on every dataset.
For Celeba, which has the highest IDE, we additionally train VAEs with latent dimensions $n=42, 52, 62, 100, 150, 200$.

\paragraph{Estimations of the ID}
For all the models, we estimate the ID of the layers' activations using 3 batches of 10,000 data examples each.
As in~\citet{Pope2021}, the MLE scores are computed with $k=3,5,10,20$.

Additional details on our implementation can be found in~\Appref{sec:xp-setup} and
our code is available\ifanonymous\else~at \url{https://github.com/bonheml/VAE_learning_dynamics}\fi.
    \section{Results}\label{sec:res}
In this section, we will analyse the results of the experiment detailed in~\Secref{sec:xp}.~First, we will review the IDE
of the different datasets in~\Secref{subsec:res-data}.
Then, in~\Secref{subsec:res-vaes}, we will discuss the variation of IDs between different layers of VAEs when we change the number of latent dimensions
and how it evolves during training.
Finally, based on the findings of these sections, we will answer the objective of~\Secref{sec:xp} by proposing FONDUE --- an algorithm
to automatically find the optimal number of latent dimensions for VAEs in an efficient and unsupervised way ---
in~\Secref{subsec:res-fondue}.

\subsection{Estimating the intrinsic dimensions of the datasets}\label{subsec:res-data}

\begin{figure}[ht!]
    \centering
    \includegraphics[width=0.35\textwidth]{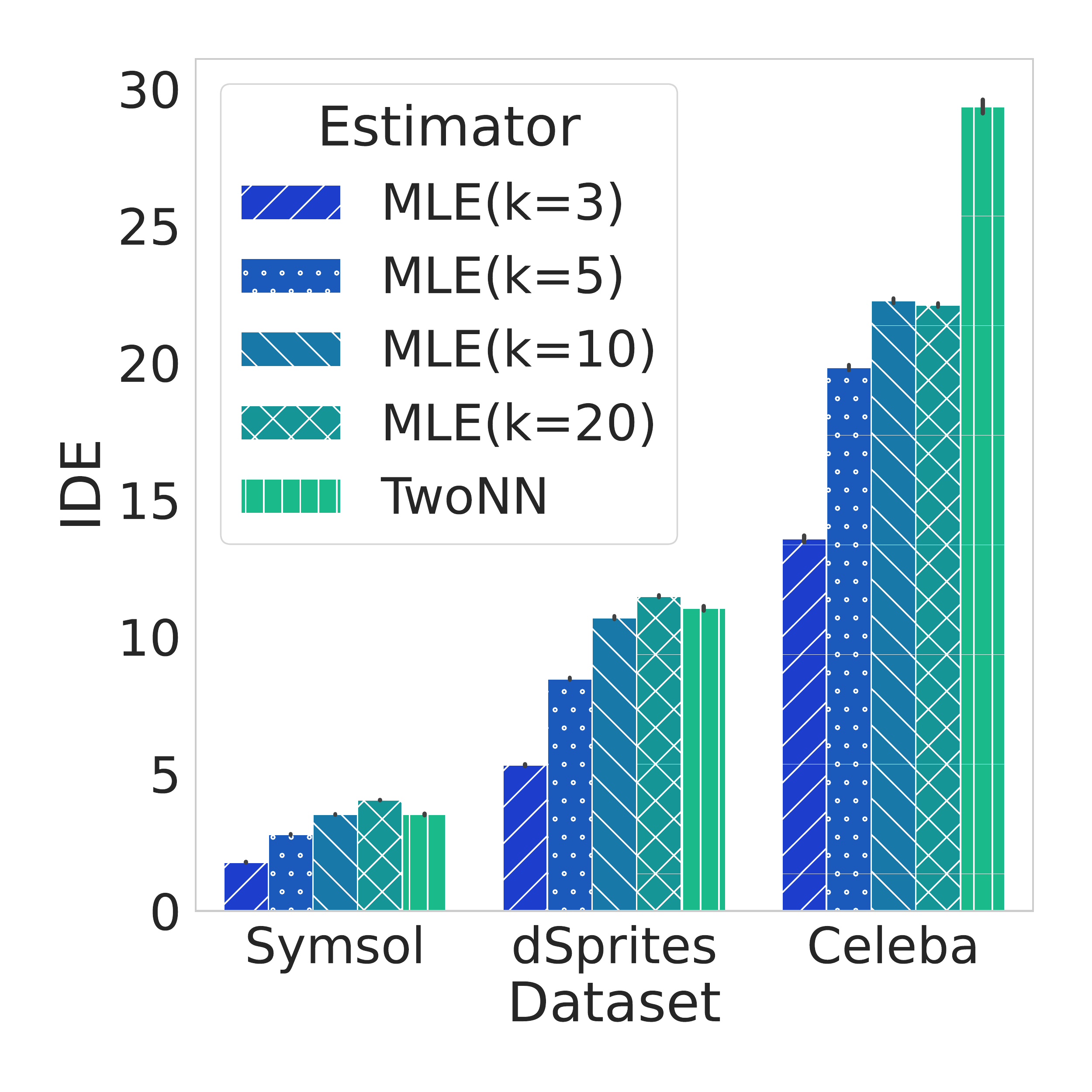}
    \caption{Intrinsic dimension estimation of dSprites, Celeba, and Symsol using different ID estimators.
    We can see that MLE have very close estimates for 10 and 20 neighbours, which generally agree with TwoNN estimates. However,
        TwoNN seems to overestimate the ID of Celeba.}
    \label{fig:ide-data}
\end{figure}

Following~\citet{Karbauskaita2011}, we will retain for our analysis the MLE estimates which are stable for the largest
number of $k$ values, as detailed in~\Secref{subsec:bg-ide}.
We can see in~\Figref{fig:ide-data} that the MLE estimations become stable when $k$ is between 10 and 20, similar to what was
reported by~\citet{Levina2004}.~These IDEs are also generally close to TwoNN estimations, except for Celeba, where TwoNN seems to
overestimate the ID, as previously reported by~\citet{Pope2021}.
In the rest of this paper, we will thus consider the IDEs obtained from MLE with $k=20$ as our most likely IDEs.

As mentioned in~\Secref{sec:xp}, we have selected 3 datasets of increasing intrinsic dimensionality:
Symsol~\citep{Murphy2021}, dSprites~\citep{Higgins2017}, and Celeba~\citep{Liu2015}.
Celeba's IDE was previously estimated to be 26 for MLE with $k=20$~\citep{Pope2021}, and we know that Symsol and dSprites
have 2 and 5 generative factors, respectively.~We thus expect their IDEs to be close to these values.
~We can see in~\Figref{fig:ide-data} that MLE and TwoNN overestimate the IDs of Symsol and dSprites, with IDEs of 4
and 11 instead of the expected 2 and 5.~Our result for Celeba is close to~\citet{Pope2021} with an estimate of 22;
the slight difference may be attributed to the difference in our averaging process (\citet{Pope2021} used
the averaging described by~\citet{Mackay2005} instead of the original averaging of~\citet{Levina2004} given in~\Eqref{eq:ide-mle-avg}).

Overall, we can see that we get an upper bound on the true ID of the data for the datasets whose ID are known.
However, as we will study the variations of ID over different layers, our experiment will not be impacted by
any overestimation of the true ID.

\subsection{Analysing the IDE of the different layers of VAEs}\label{subsec:res-vaes}
Now that we have an IDE for each of the datasets, we are interested in observing how the ID of VAEs varies between layers
and when they are trained with different numbers of latent variables.

\begin{figure}[ht!]
    \centering
    \begin{subfigure}{.45\textwidth}
        \centering
        \includegraphics[width=\linewidth]{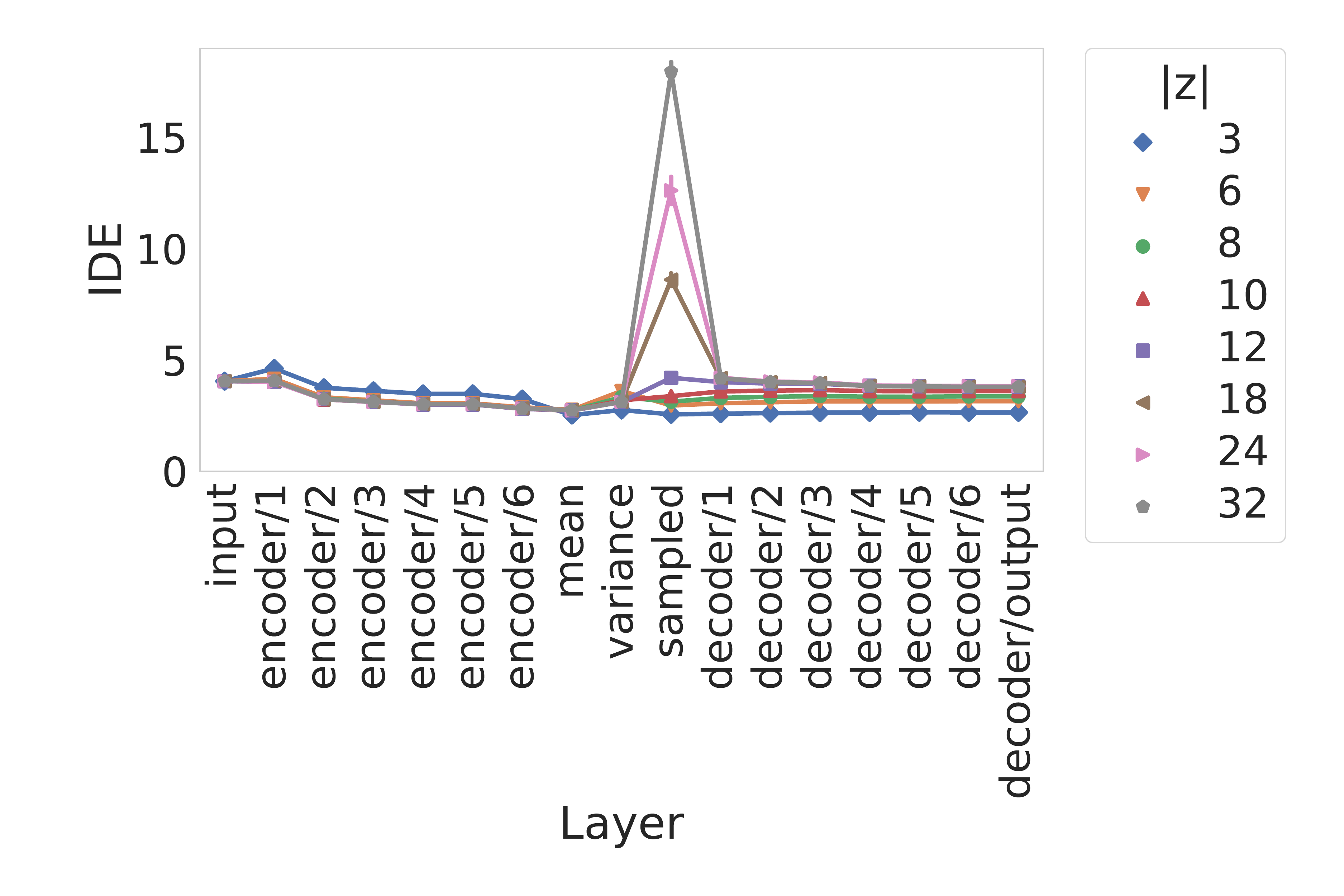}
        \caption{Symsol}
        \label{fig:ide-layers-symsol}
    \end{subfigure}%
    \begin{subfigure}{.45\textwidth}
        \centering
        \includegraphics[width=\linewidth]{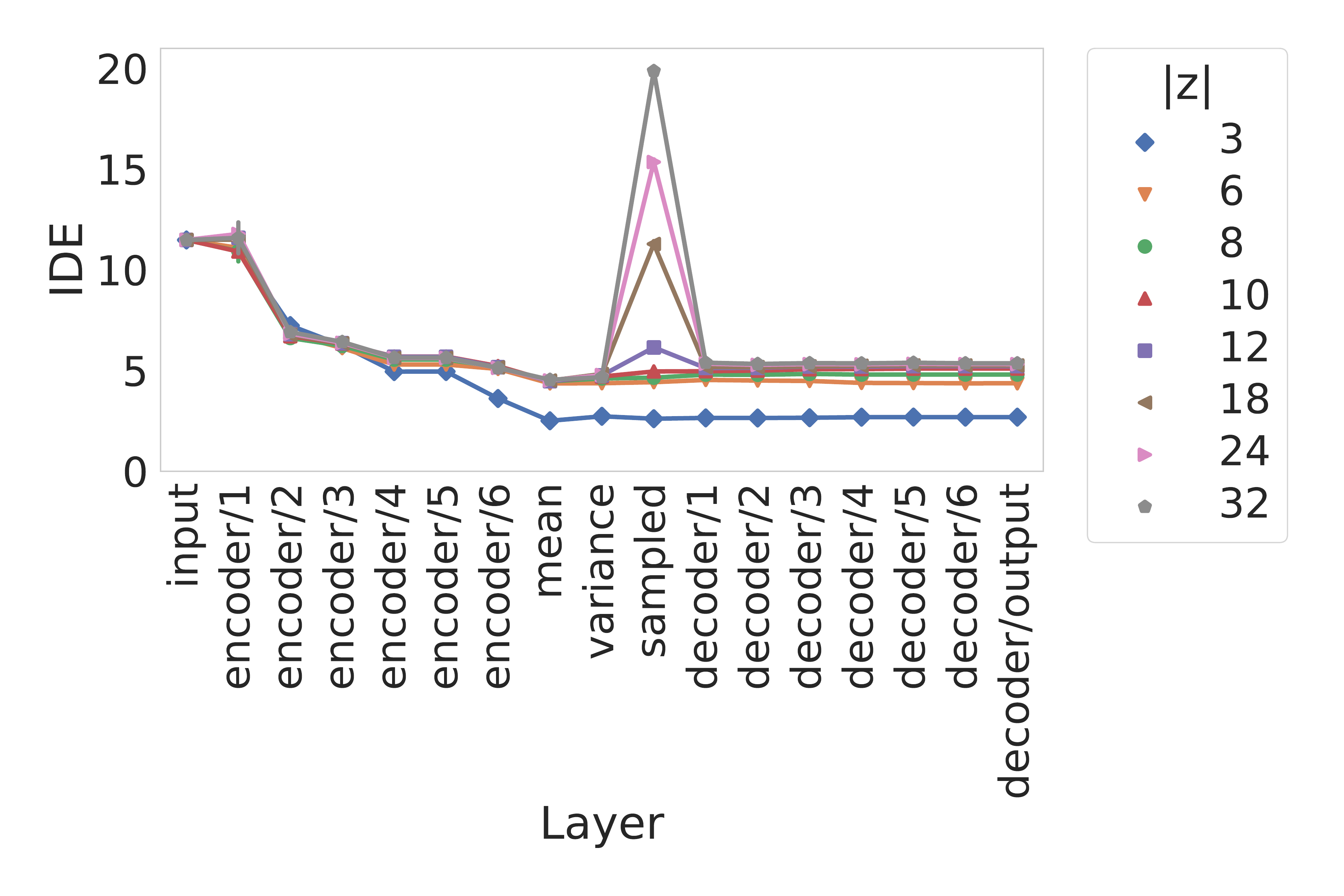}
        \caption{dSprites}
        \label{fig:ide-layers-dsprites}
    \end{subfigure}\\
    \begin{subfigure}{.45\textwidth}
        \centering
        \includegraphics[width=\linewidth]{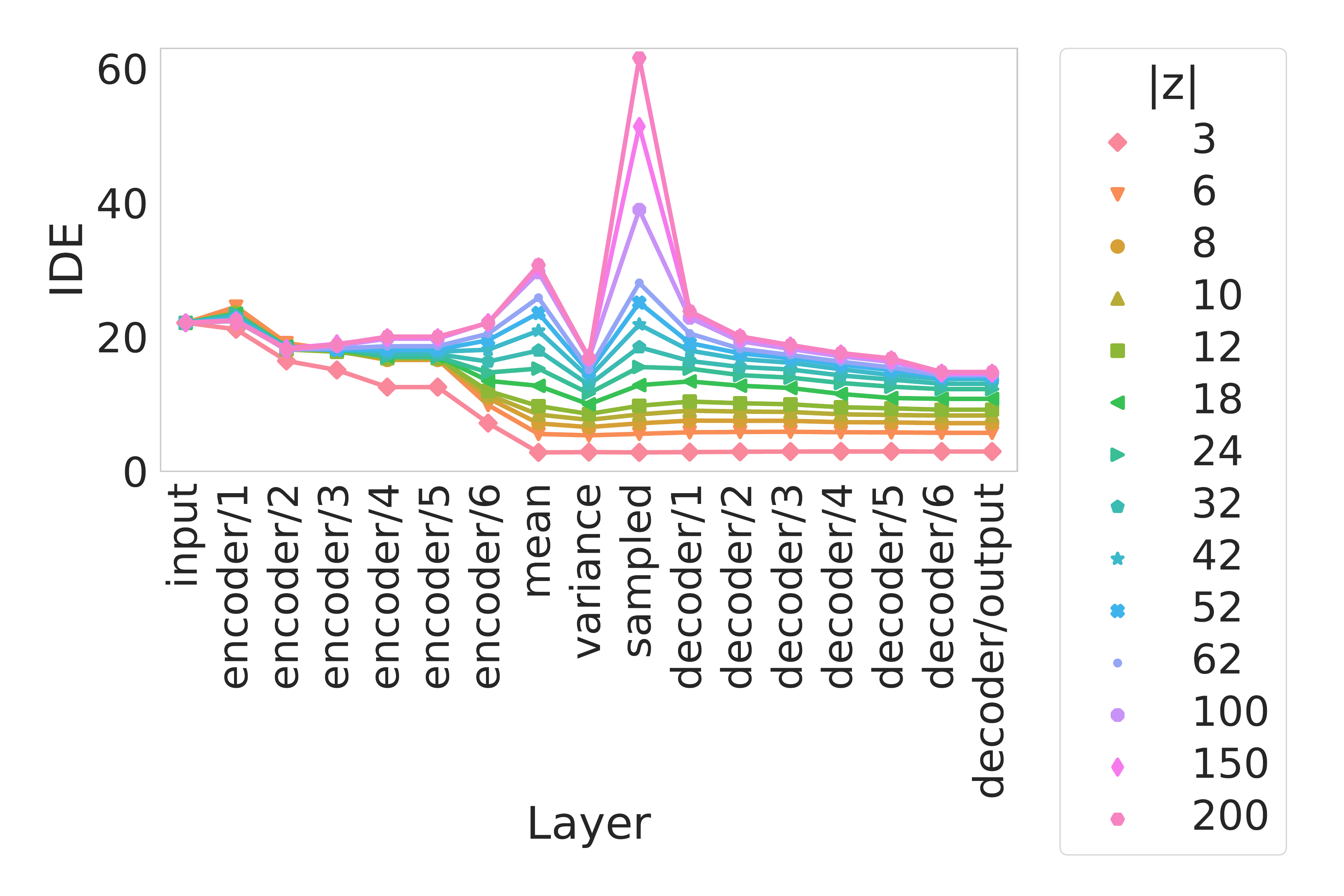}
        \caption{Celeba}
        \label{fig:ide-layers-celeba}
    \end{subfigure}%
    \begin{subfigure}{.45\textwidth}
            \centering
        \includegraphics[width=\textwidth]{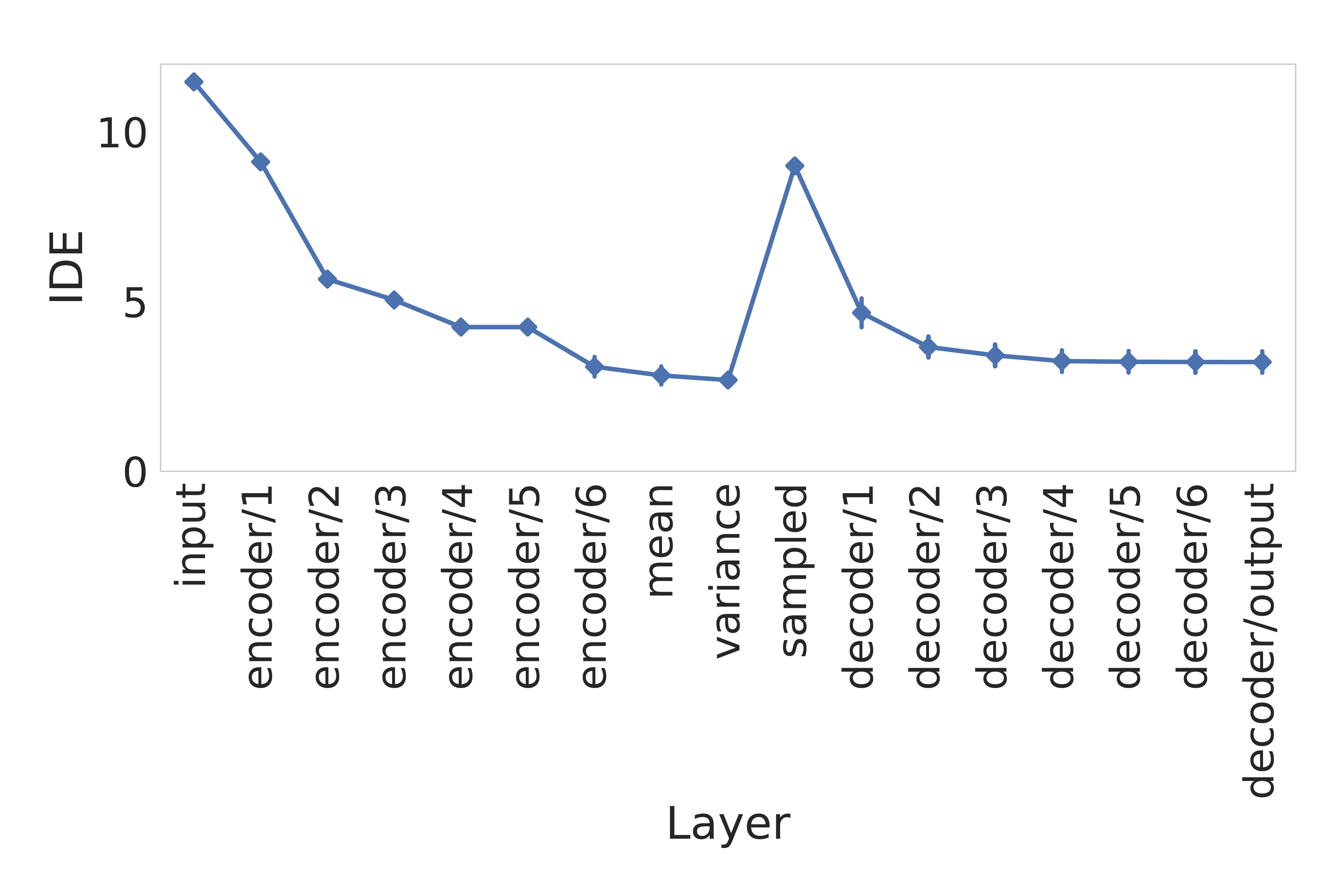}
        \caption{$\beta$-VAE trained on dSprites with 10 latent variables and $\beta=20$.}
        \label{fig:ide-collapse}
    \end{subfigure}
    \caption{Intrinsic dimension estimation of VAEs trained with an increasing number of latent dimensions $|\rvz|$.
    (a), (b), and (c) show the results on Symsol, dSprites, and Celeba, respectively.
    (d) shows the results of $\beta$-VAEs trained on dSprites with 10 latent variables and $\beta=20$ to cause posterior collapse.}
    \label{fig:ide-layers}
\end{figure}

\paragraph{The ID of the encoder decreases, but the ID of the decoder stays constant}
We can see in~\Figref{fig:ide-layers} that the ID of the representations learned by the encoder decreases
until we reach the mean and variance layers, which is consistent with the observations reported for
classification~\citep{Ansuini2019}. Interestingly, for dSprites and Symsol, when the number of latent variables is at
least equal to the ID of the data, the IDE of the mean and variance representations is very close to the true data ID.
After a local increase of the ID in the sampled representations, the ID of the decoder representations stays close to
the ID of the mean representations and does not change much between layers.

\begin{figure}[ht!]
    \centering
    \begin{subfigure}{.3\textwidth}
        \centering
        \includegraphics[width=\linewidth]{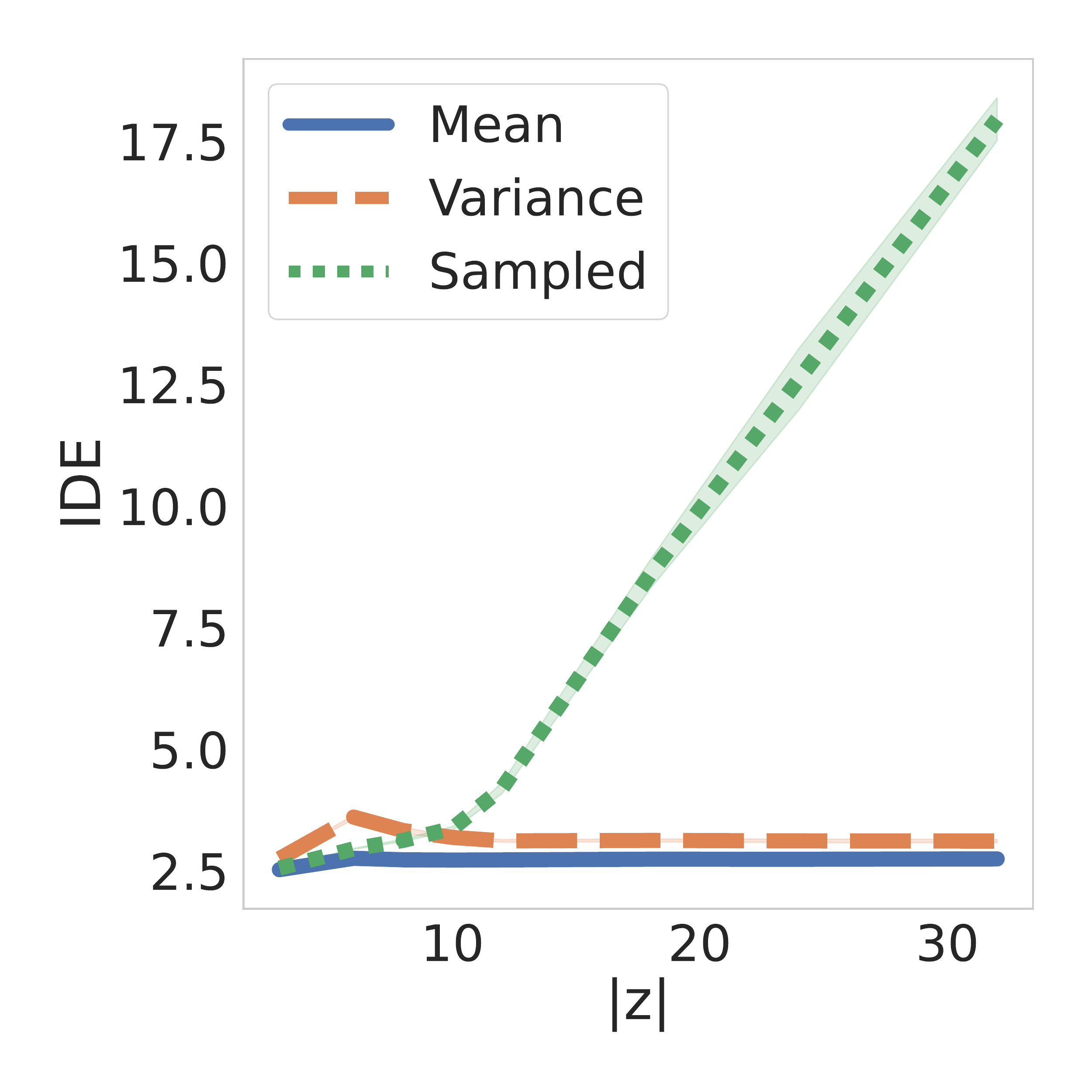}
        \caption{Symsol}
        \label{fig:ide-latents-symsol}
    \end{subfigure}%
    \begin{subfigure}{.3\textwidth}
        \centering
        \includegraphics[width=\linewidth]{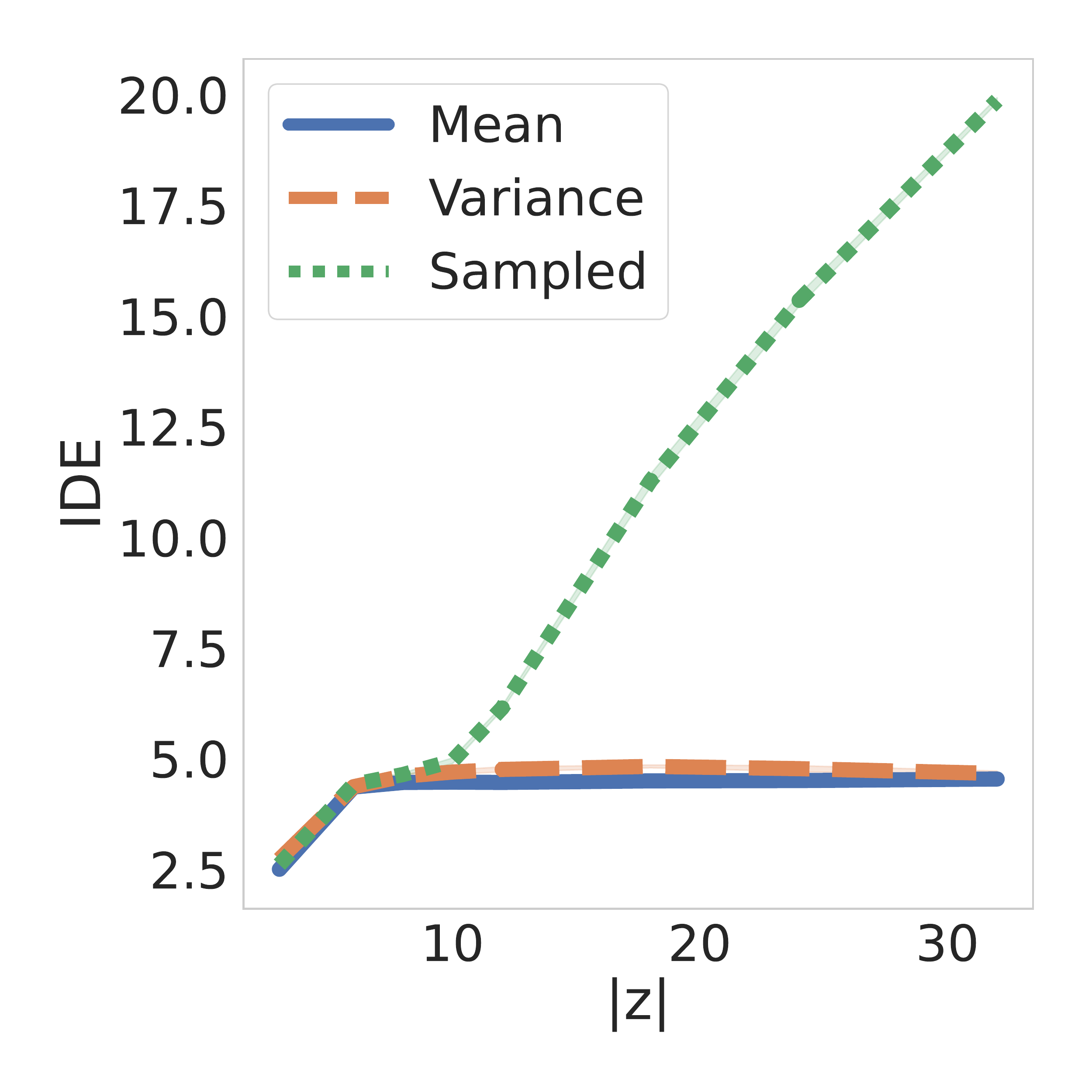}
        \caption{dSprites}
        \label{fig:ide-latents-dsprites}
    \end{subfigure}%
    \begin{subfigure}{.3\textwidth}
        \centering
        \includegraphics[width=\linewidth]{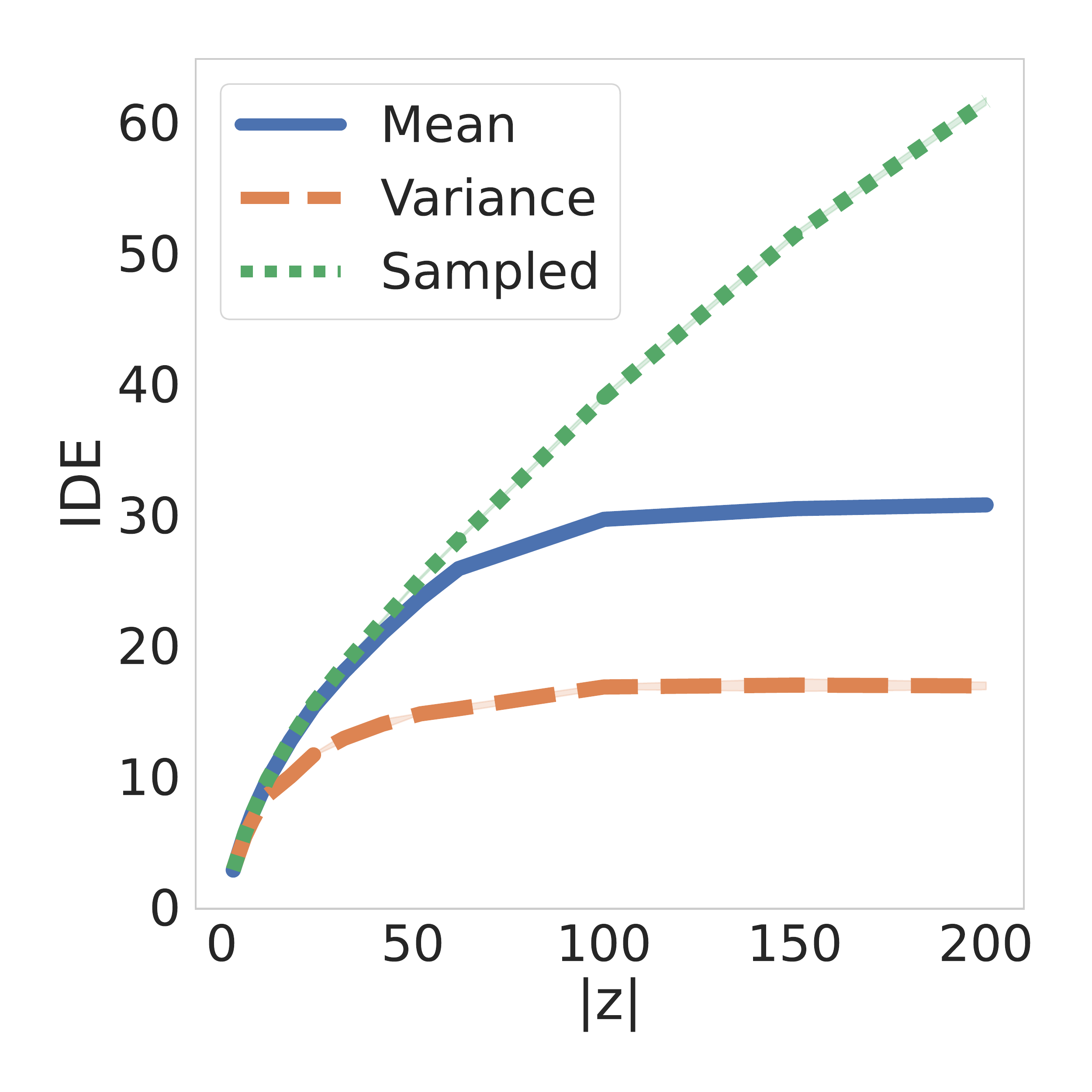}
        \caption{Celeba}
        \label{fig:ide-latents-celeba}
    \end{subfigure}
    \caption{Intrinsic dimension estimation of the mean, variance, and sampled representations of VAEs trained with an
    increasing number of latent dimensions $|\rvz|$.
        (a), (b), and (c) shows the results on Symsol, dSprites, and Celeba, respectively.}
    \label{fig:ide-latents}
\end{figure}

\begin{figure}[ht!]
    \centering
    \begin{subfigure}{.3\textwidth}
        \centering
        \includegraphics[width=\linewidth]{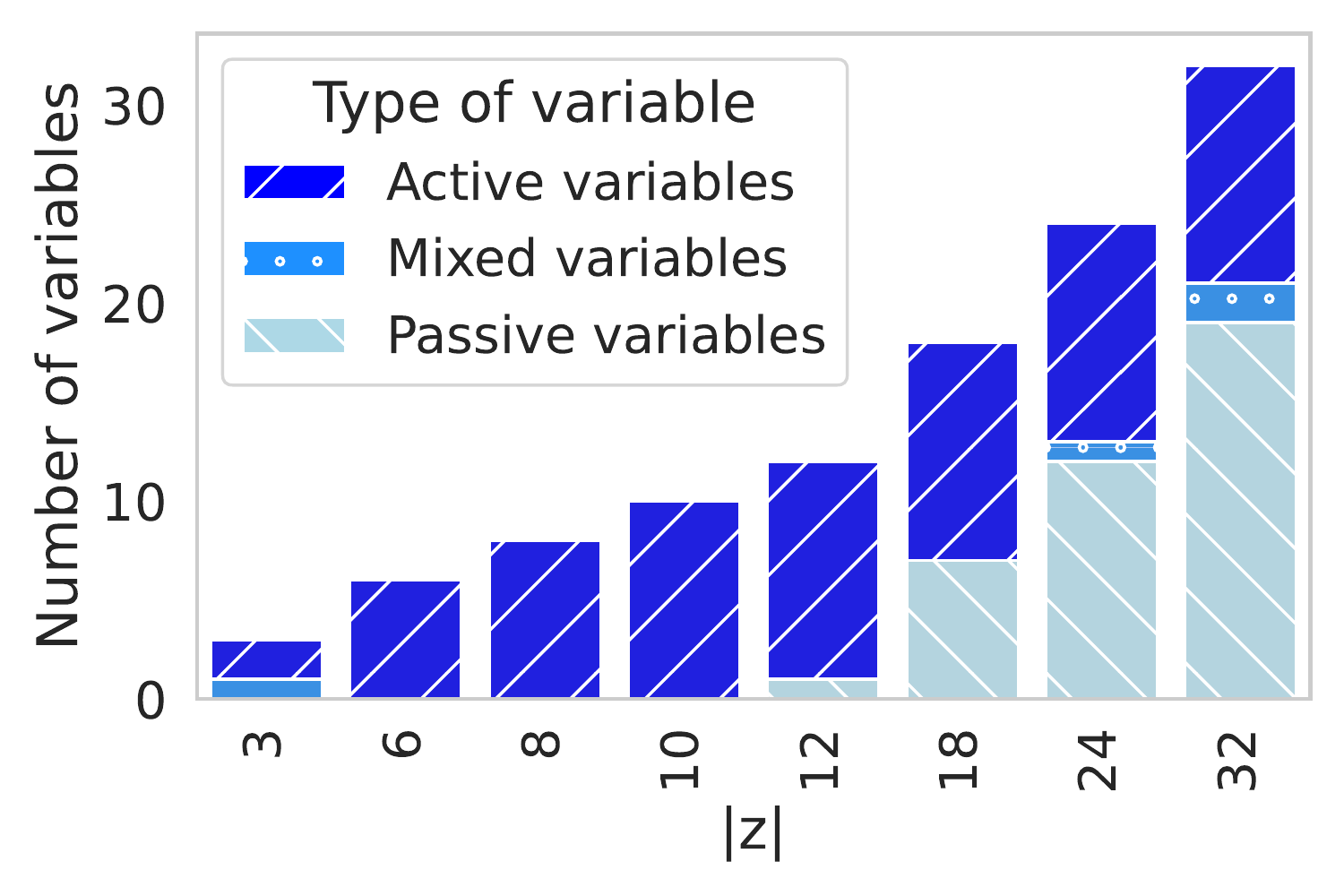}
        \caption{Symsol}
        \label{fig:var-type-symsol}
    \end{subfigure}%
    \begin{subfigure}{.3\textwidth}
        \centering
        \includegraphics[width=\linewidth]{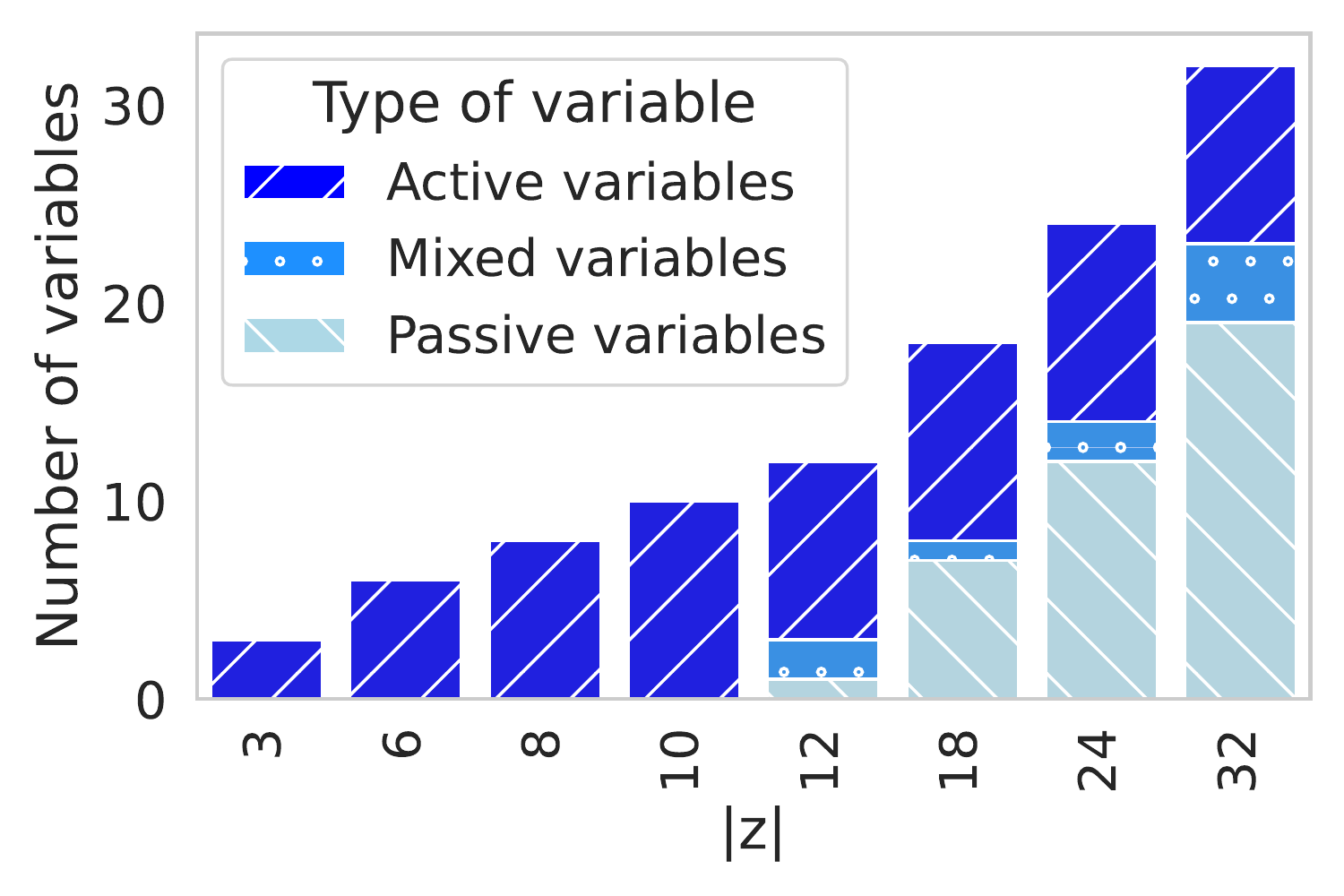}
        \caption{dSprites}
        \label{fig:var-type-dsprites}
    \end{subfigure}%
    \begin{subfigure}{.3\textwidth}
        \centering
        \includegraphics[width=\linewidth]{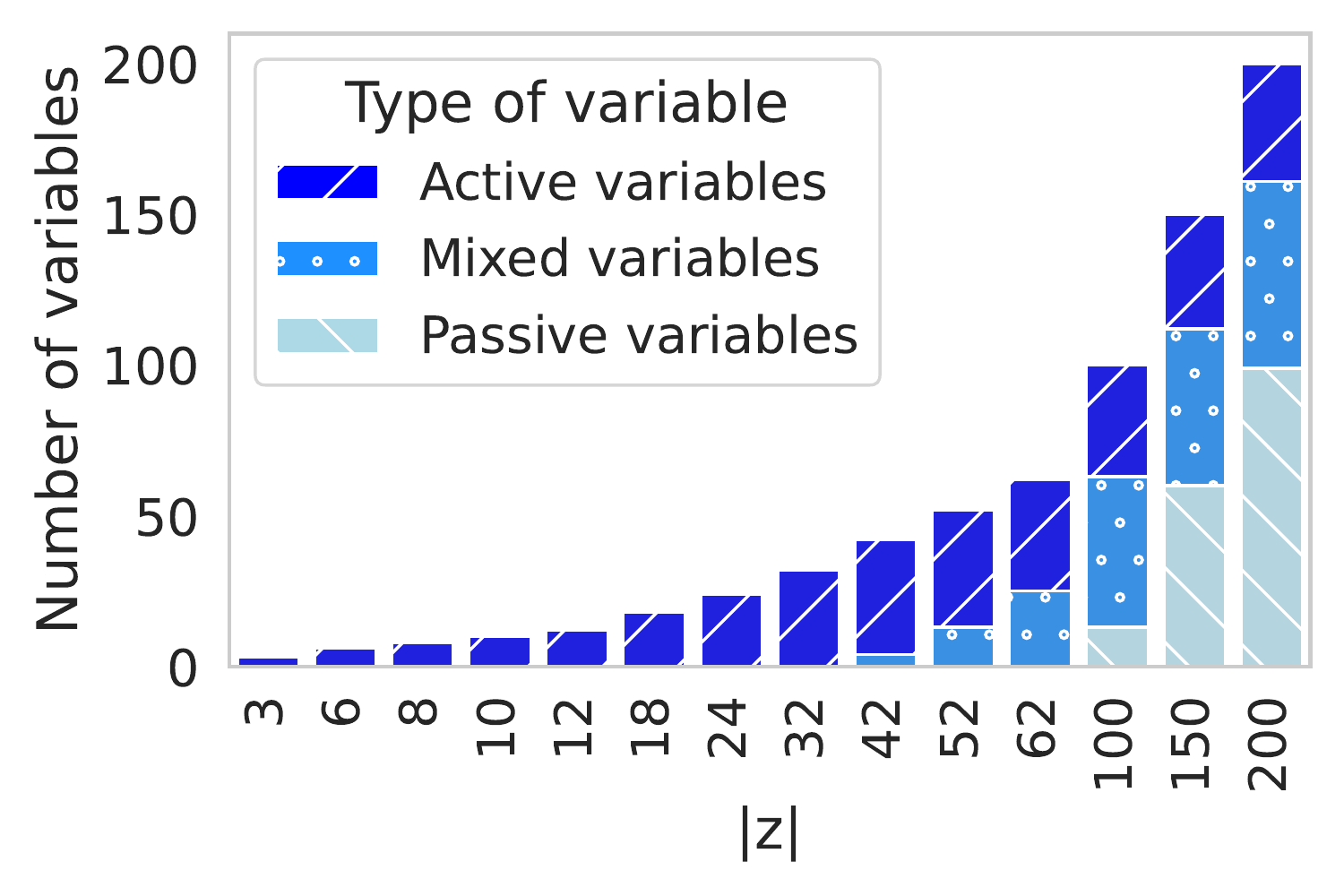}
        \caption{Celeba}
        \label{fig:var-type-celeba}
    \end{subfigure}
    \caption{Quantity of active, mixed, and passive variables of VAEs trained with an increasing number of latent dimensions $|\rvz|$.
        (a), (b), and (c) show the results on Symsol,
        dSprites, and Celeba, respectively.}
    \label{fig:var-type}
\end{figure}

\paragraph{Mean and sampled representations have different IDEs}
Looking into the IDEs of mean and sampled representations in~\Figref{fig:ide-latents}, we see
a clear pattern emerge: when increasing the number of latent variables the IDEs remain similar up to a point,
then abruptly diverge.
As discussed in~\Secref{subsec:bg-VAEs}, once a VAE has enough latent variables to encode the information needed by the decoder, the remaining
variables will become passive to minimise the KL divergence in~\Eqref{eq:beta-vae}.
This phenomenon will naturally occur when we increase the number of latent variables.
~\citet{Bonheme2021} observed that, in the context of the polarised
regime, passive variables were very different in mean and sampled representations.
Indeed, for sampled representations, the set of passive variables will be sampled from $\N(0, I)$ where they will stay close to 0
with very low variance in mean representations. They also introduced the concept of mixed variables --- variables
that are passive only for some data examples --- and shown that they were also leading to different mean and sampled
representations, albeit to a minor extent.
We can thus hypothesise that the difference between the mean and sampled IDEs grows with the number of mixed and passive variables.
This is verified by computing the number of active, mixed, and passive variables using the method of~\citet{Bonheme2021}, as shown in~\Figref{fig:var-type}.

\paragraph{What happens in the case of posterior collapse?} By using a $\beta$-VAE with very large $\beta$ (e.g., $\beta = 20$),
one can induce posterior collapse, where a majority of the latent variables become passive and prevent the
decoder from accessing sufficient information about the input to provide a good reconstruction.
This phenomenon is illustrated in~\Figref{fig:ide-collapse}, where the IDs of the encoder are similar to what one would obtain for
a well performing model in the first 5 layers, indicating that these early layers of the encoder still encode some useful information about the data.
The IDs then drop in the last three layers of the encoder, indicating that most variables are passive, and only a very small amount of information is retained.
The ID of the sampled representation (see \textit{sampled} in ~\Figref{fig:ide-collapse}) is then artificially inflated by the passive variables and becomes very close to the number of dimensions $|\rvz|$.
From this, the decoder is unable to learn much and has thus a low ID, close to the ID of the mean representation (see the points on the RHS of~\Figref{fig:ide-collapse}).

\begin{figure}[ht!]
    \centering
    \includegraphics[width=0.5\linewidth]{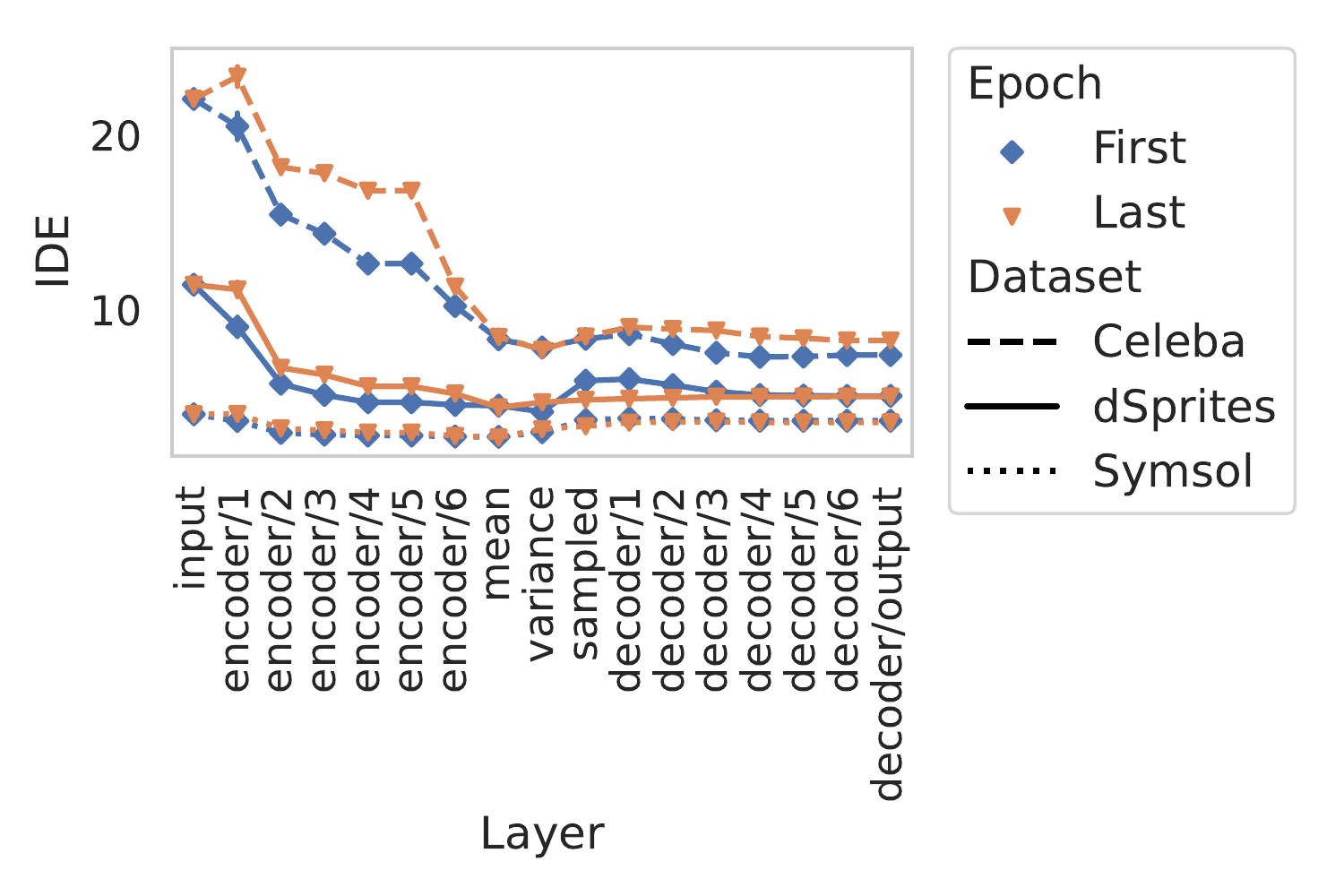}
    \caption{The evolution over multiple epochs of the IDE of the representations learned by VAEs using 10 latent variables
    on Symsol, dSprites, and Celeba.}
    \label{fig:epoch}
\end{figure}

\paragraph{The IDs of the model's representations do not change much after the first epoch}
The IDs of the different layers do not change much after the first epoch for
well-performing models (see~\Figref{fig:epoch}).
However, for Celeba, whose number of latent dimensions is lower than the data IDE and thus cannot
reconstruct the data well, the ID tends to change more in the early layers of the encoder, with a higher variance in IDE.

\subsection{Finding the optimal number of dimensions by unsupervised estimation}\label{subsec:res-fondue}
As discussed in~\Secref{subsec:res-vaes}, the IDs of the mean and sampled representations start to diverge when
(unused) passive variables appear, and this is already visible after the first epochs of training.
We can thus use the difference of IDs between the mean and sampled representations to find the
number of latent dimensions retaining the most information while remaining sufficiently compressed (i.e., no passive variables).
To this aim, we propose an algorithm for Finding the Optimal Number of Dimensions from Unsupervised Estimation (FONDUE).
\begin{thm}\label{thm:fondue}
Any execution of FONDUE (\Algref{alg:fondue}) returns the largest number of dimensions $p$ for which $IDE_z - IDE_{\mu} \leqslant threshold$,
where $IDE_z$, and $IDE_{\mu}$ are the IDEs of the sampled and mean representations, respectively.
\end{thm}

\begin{proof}[Proof Sketch]
In~\Algref{alg:fondue}, we define a lower and upper bound of the ID estimate, $l$ and $u$, and update
the predicted number of latent variables $p$ until, after $i$ iterations, $p_i=l_i$.
Using the loop invariant $l_i \leqslant p_i \leqslant u_i$, we can show that the algorithm
terminates when $l_i = p_i = \operatorname{floor}\left(\frac{l_i + u_i}{2}\right)$, which can only be reached when $u_i = p_i + 1$, that
is, when $p_i$ is the maximum number of latent dimensions for which we have $IDE_z - IDE_{\mu} \leqslant threshold$.
See~\Appref{sec:app-fondue-proof} for the full proof.
\end{proof}
To ensure stable ID estimates, we computed FONDUE multiple times, gradually increasing the number of
epochs $e$ until the predicted $p$ stopped changing. As reported in~\Tableref{table:fondue-ides}, the results were already
stable after one epoch, except for Symsol which needed two\footnote{Note that the numbers of epochs given
in~\Tableref{table:fondue-ides} correspond to the minimum number of epochs needed for FONDUE to be stable.
For example, if we obtain the same score after 1 and 2 epochs, the number of epochs given in~\Tableref{table:fondue-ides} is 1.}.
We set a fixed threshold $t=0.2$ (20\% of the data IDE) in all our experiments and
used memoisation (see~\Algref{alg:get-mem}) to avoid unnecessary retraining and speed up~\Algref{alg:fondue}.

\noindent\begin{minipage}{0.65\textwidth}
             \captionof{algorithm}{FONDUE}\label{alg:fondue}
             \begin{algorithmic}[1]
                 \Procedure{FONDUE}{$t, IDE_{data}, epochs$}
                     \State $l \gets 0$ \Comment{Lower bound}
                     \State $u \gets \infty$ \Comment{Upper bound}
                     \State $p \gets IDE_{data}$ \Comment{Current number of latent dimensions}
                     \State $mem \gets \{\}$
                     \State $threshold \gets \frac{t \times p}{100}$
                     \While{$p \neq l$}
                         \State $IDE_z, IDE_{\mu} \gets \operatorname{GET-MEM}(mem, p, epochs)$
                         \If{$IDE_z - IDE_{\mu} \leqslant threshold$} \Comment{\Figref{fig:fondue-1}}
                            \State $l \gets p$
                            \State $p \gets \min(p \times 2, u)$
                         \Else \Comment{\Figref{fig:fondue-2}}
                            \State $u \gets p$
                            \State $p \gets \operatorname{floor}\left(\frac{l + u}{2}\right)$
                         \EndIf
                     \EndWhile
                     \State \textbf{return} $p$
                 \EndProcedure
             \end{algorithmic}
             \hfill
             \captionof{algorithm}{GET-MEM}\label{alg:get-mem}
             \begin{algorithmic}[1]
                 \Procedure{GET-MEM}{$mem, p, e$}
                         \If{$mem[p] = \emptyset$}
                            \State $vae \gets \operatorname{TRAIN-VAE}(dim=p, n\_epochs=e)$
                            \State $IDE_z, IDE_{\mu} \gets IDEs(vae)$
                            \State $mem[p] \gets IDE_z, IDE_{\mu}$
                         \EndIf
                         \State \textbf{return} $mem[p]$
                 \EndProcedure
             \end{algorithmic}
\end{minipage}%
\hfill
\begin{minipage}[t]{.3\textwidth}
    \centering
    \includegraphics[height=0.15\textheight]{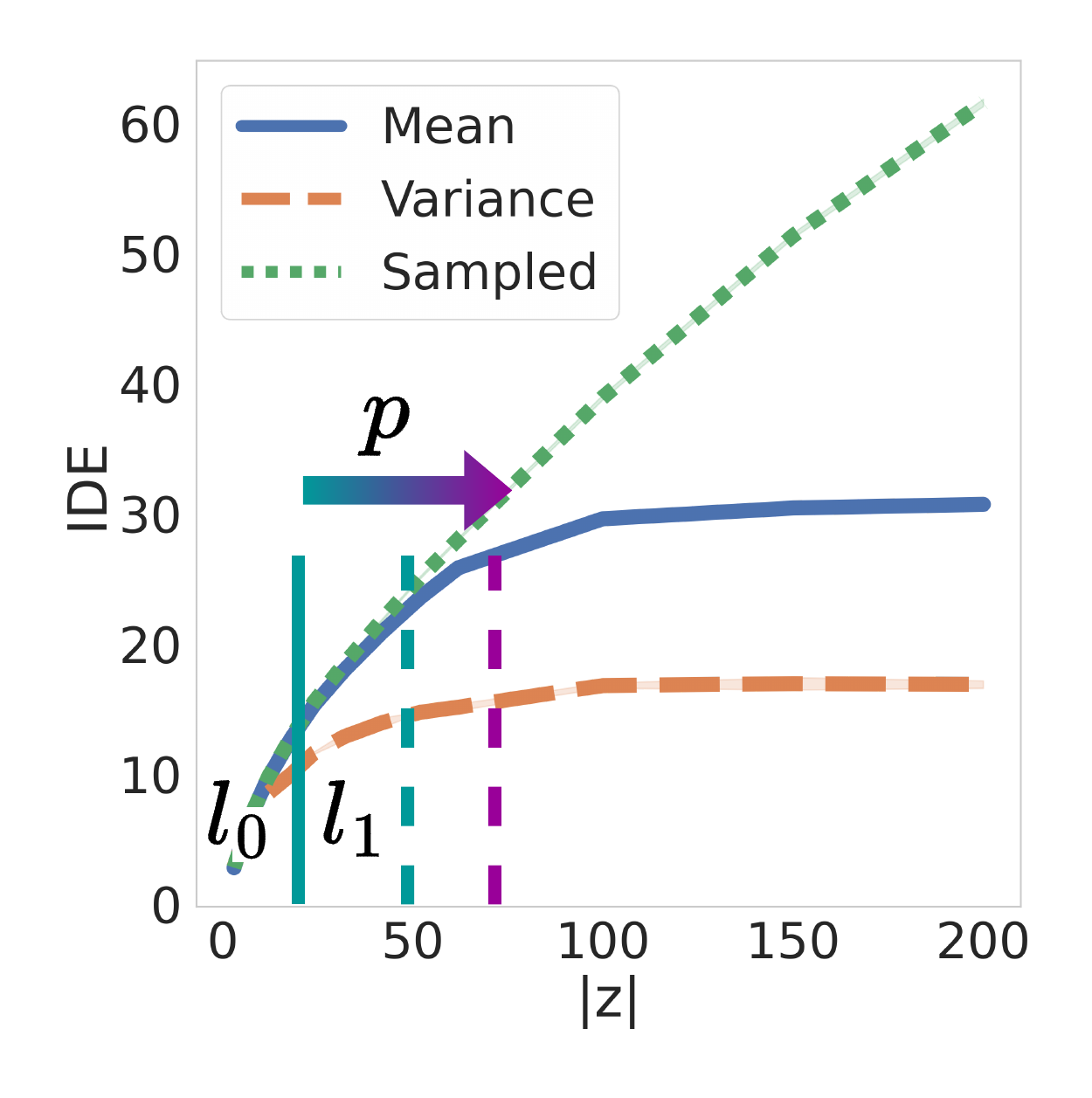}
    \captionof{figure}{Update $l$ and increase $p$ until $IDE_z - IDE_{\mu} > threshold$.}\label{fig:fondue-1}
    \includegraphics[height=0.15\textheight]{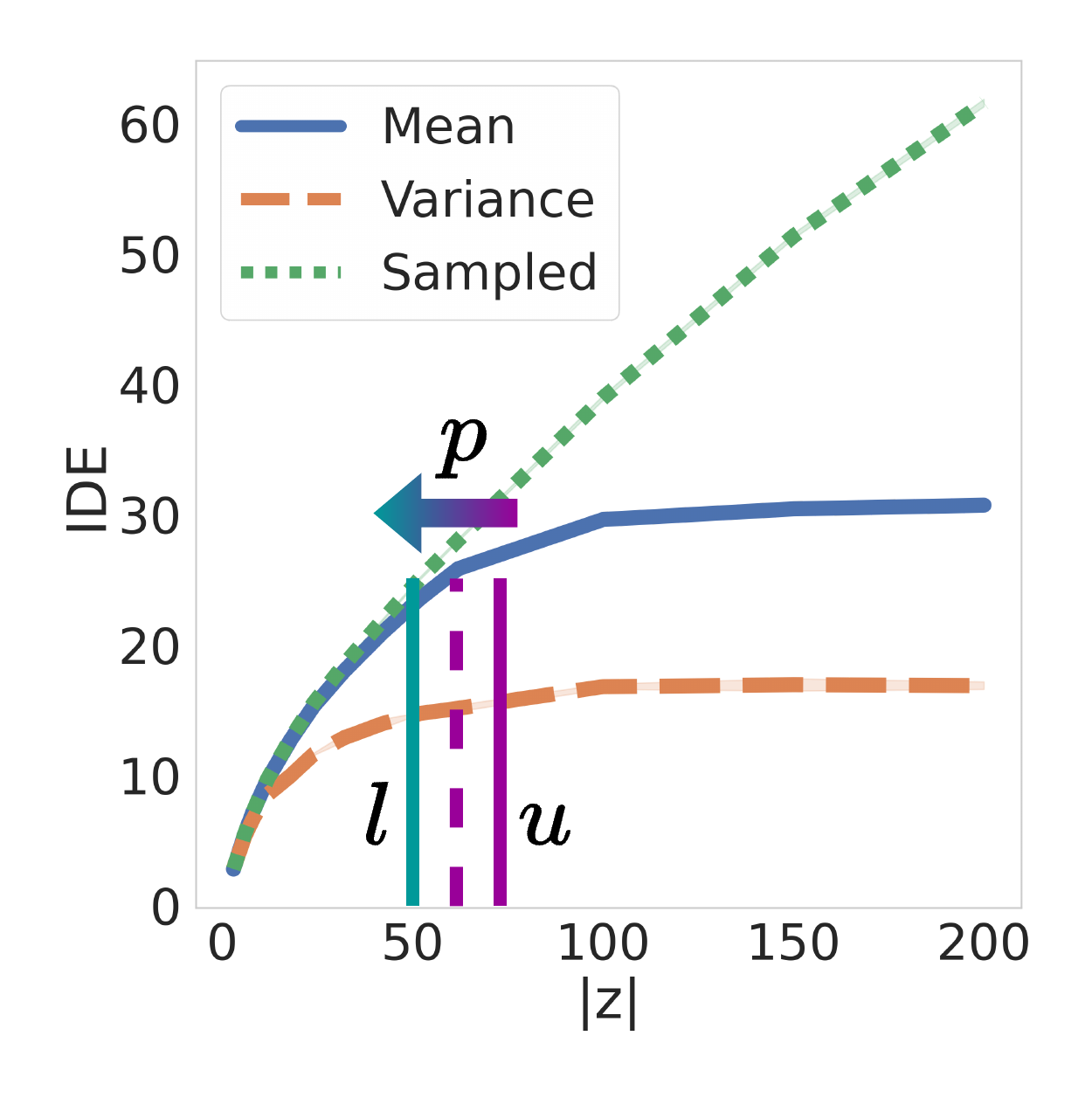}
    \captionof{figure}{Update $u$ and decrease $p$ until  $IDE_z - IDE_{\mu} \leqslant threshold$.}\label{fig:fondue-2}
\end{minipage}

\begin{figure}[ht!]
    \centering
    \begin{subfigure}{.33\textwidth}
        \centering
        \includegraphics[width=\linewidth]{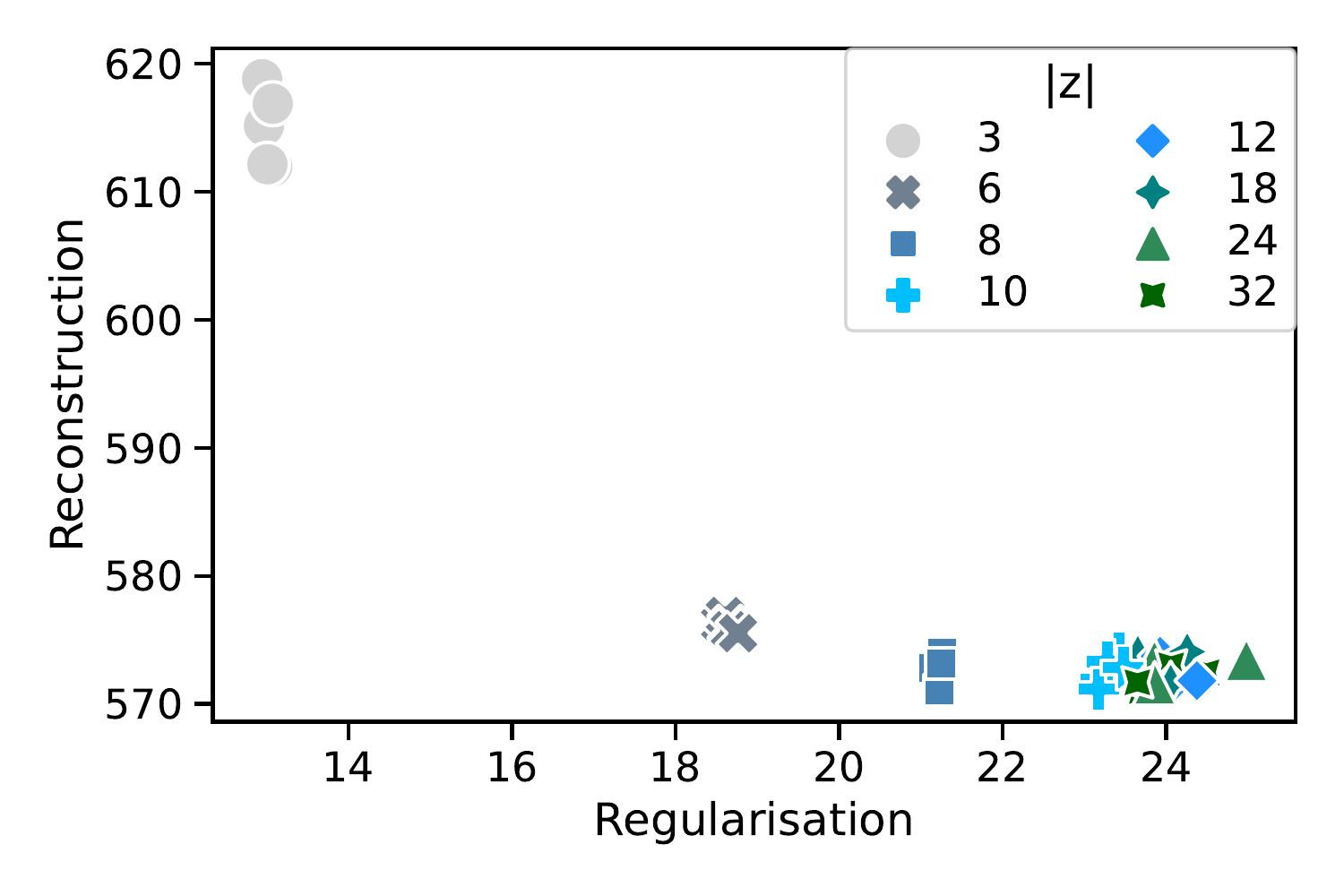}
        \caption{Symsol}
        \label{fig:fondue-rec-symsol}
    \end{subfigure}%
    \begin{subfigure}{.33\textwidth}
        \centering
        \includegraphics[width=\linewidth]{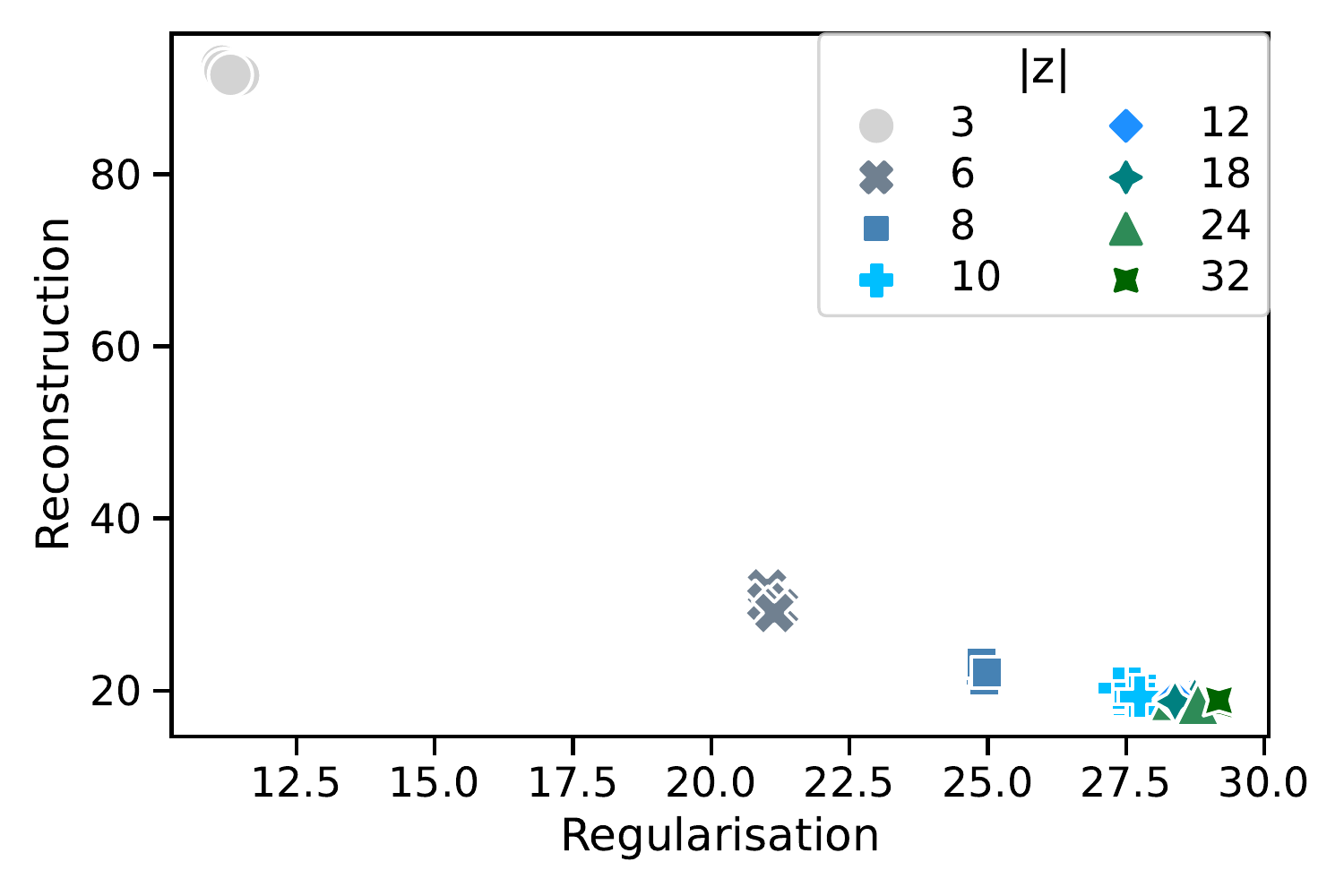}
        \caption{dSprites}
        \label{fig:fondue-rec-dsprites}
    \end{subfigure}%
    \begin{subfigure}{.33\textwidth}
        \centering
        \includegraphics[width=\linewidth]{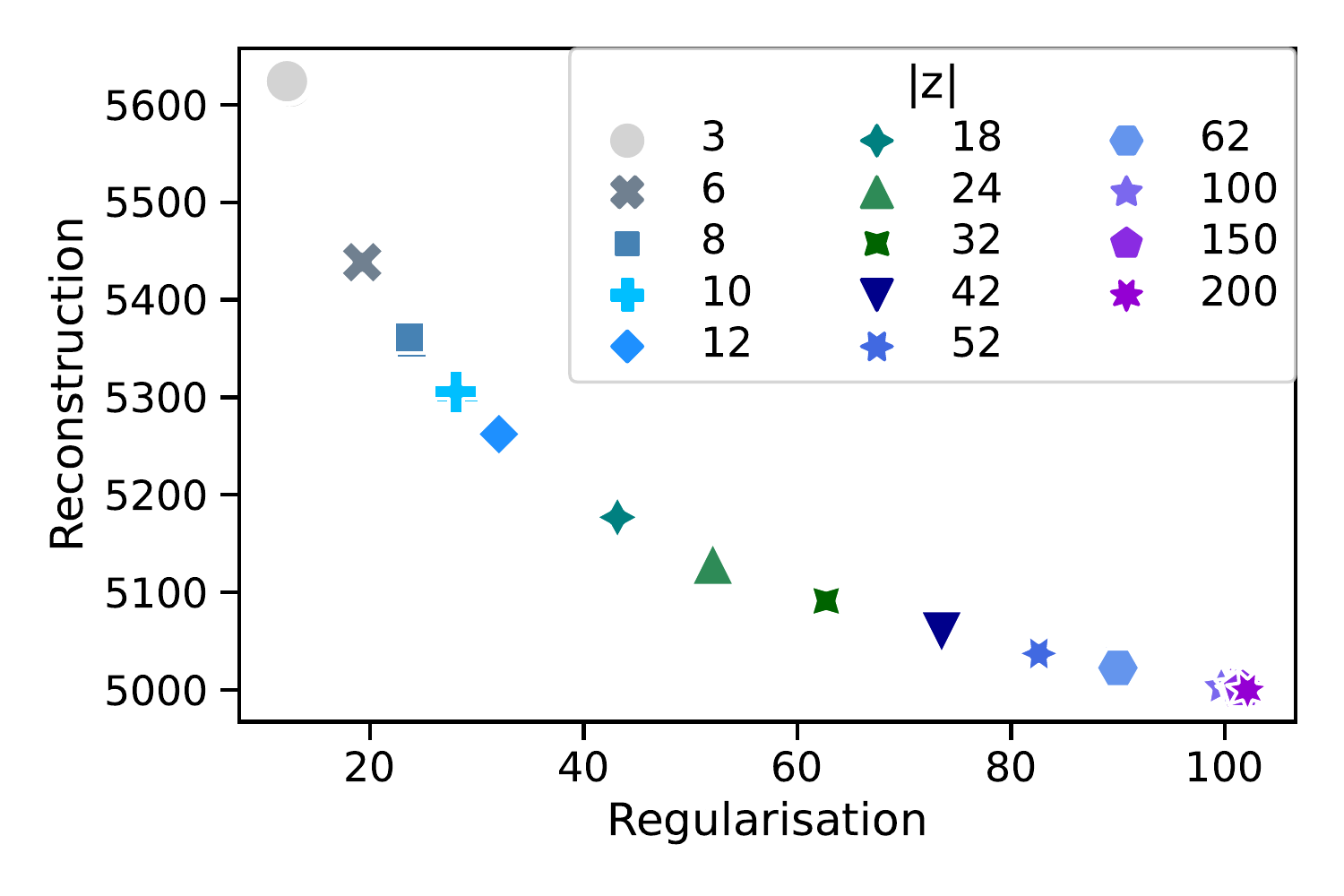}
        \caption{Celeba}
        \label{fig:fondue-rec-celeba}
    \end{subfigure}
    \caption{Reconstruction and regularisation loss of VAEs for Symsol, dSprites, and Celeba with an increasing number of latent variables.}
    \label{fig:fondue-rec}
\end{figure}

\begin{table}[ht!]
    \centering
    \caption{Number of latent variables $|\rvz|$ obtained with FONDUE. The results are averaged over 5 seeds, and computation times are reported for NVIDIA A100 GPUs.
    The computation time is given for one run of FONDUE over the minimum number of epochs needed to obtain a stable score.}
    \label{table:fondue-ides}
    \begin{tabular}{ l l l l l}
        \hline
        Dataset & Dimensionality (avg $\pm$ SD) & Time/run & Models trained & Epochs/training\\
        \hline
        \rule{0pt}{2.6ex}Symsol & 11 $\pm$ 0 & 7 min & 6 & 2\\
        dSprites & 12.2 $\pm$ 0.4 & 20 min & 5 & 1\\
        Celeba & 50.2 $\pm$ 0.9 & 14 min & 9 & 1\\
        \hline
    \end{tabular}
\end{table}

\paragraph{Analysing the results of FONDUE}
As shown in~\Tableref{table:fondue-ides}, the execution time for finding the optimal number of dimensions of a
dataset is much shorter than for fully training one model (approximately 2h using the same GPUs), making the algorithm clearly
more efficient than grid search. Moreover, the number of latent dimensions predicted by FONDUE is consistent with well-performing models.
Indeed, for dSprites and Symsol, the selected numbers of dimensions correspond to the number of dimensions after which the
reconstruction error stops decreasing and the regularisation loss remains stable (see~\Figref{fig:fondue-rec}).
For Celeba, the reconstruction loss continues to improve slightly after 50 latent dimensions, due to the addition of
mixed variables between 52 and 100 latent dimensions.~While one could increase the threshold of FONDUE to take more mixed
variables into account, this may not always be desirable. Indeed, mixed variables encode features specific to
a subtype of data examples~\citep{Bonheme2021} and will provide less compact representations for a gain in reconstruction quality which may only be marginal.

\begin{figure}[ht!]
    \centering
    \begin{subfigure}{.33\textwidth}
        \centering
        \includegraphics[width=\linewidth]{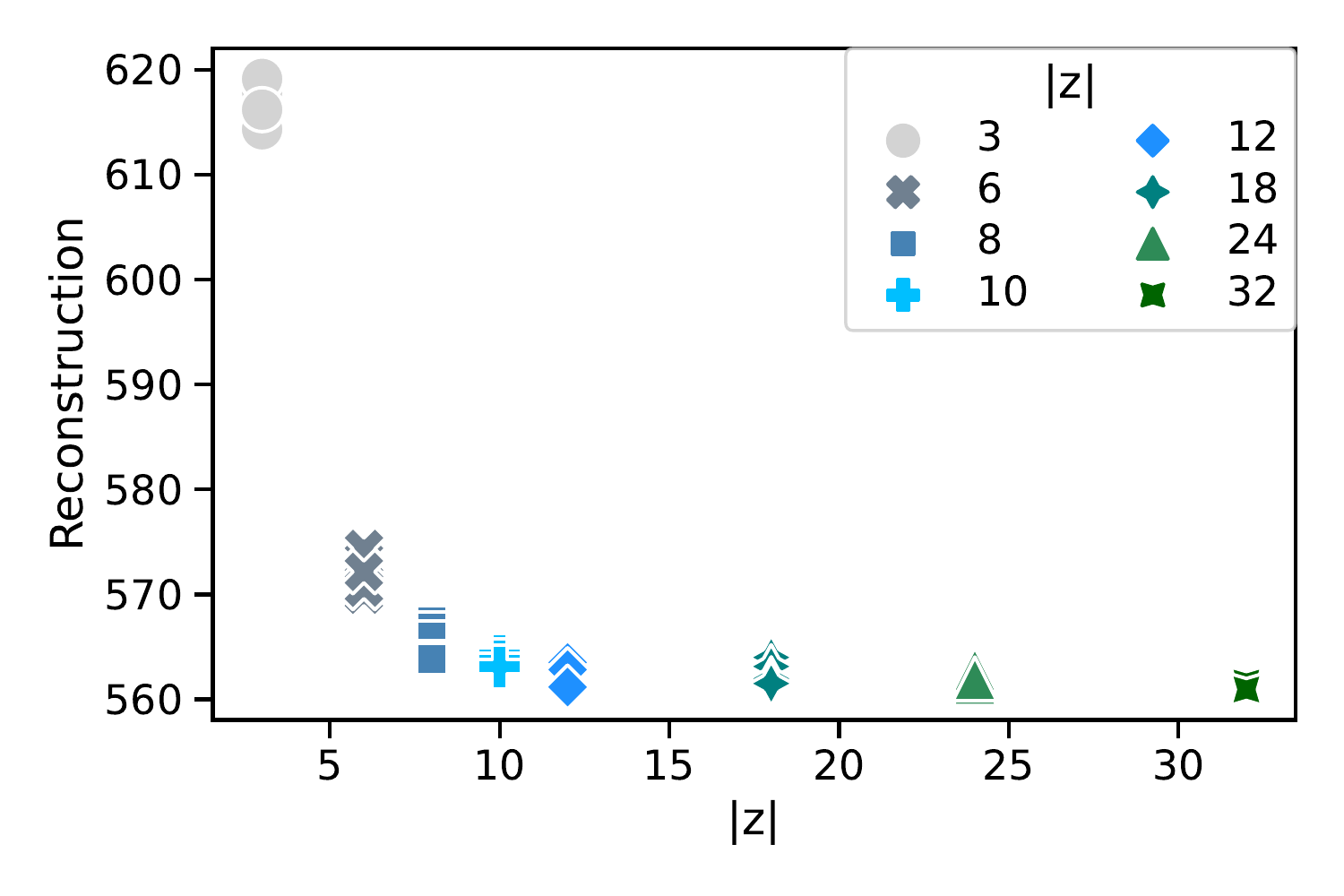}
        \caption{Symsol}
        \label{fig:symsol-ae}
    \end{subfigure}%
    \begin{subfigure}{.33\textwidth}
        \centering
        \includegraphics[width=\linewidth]{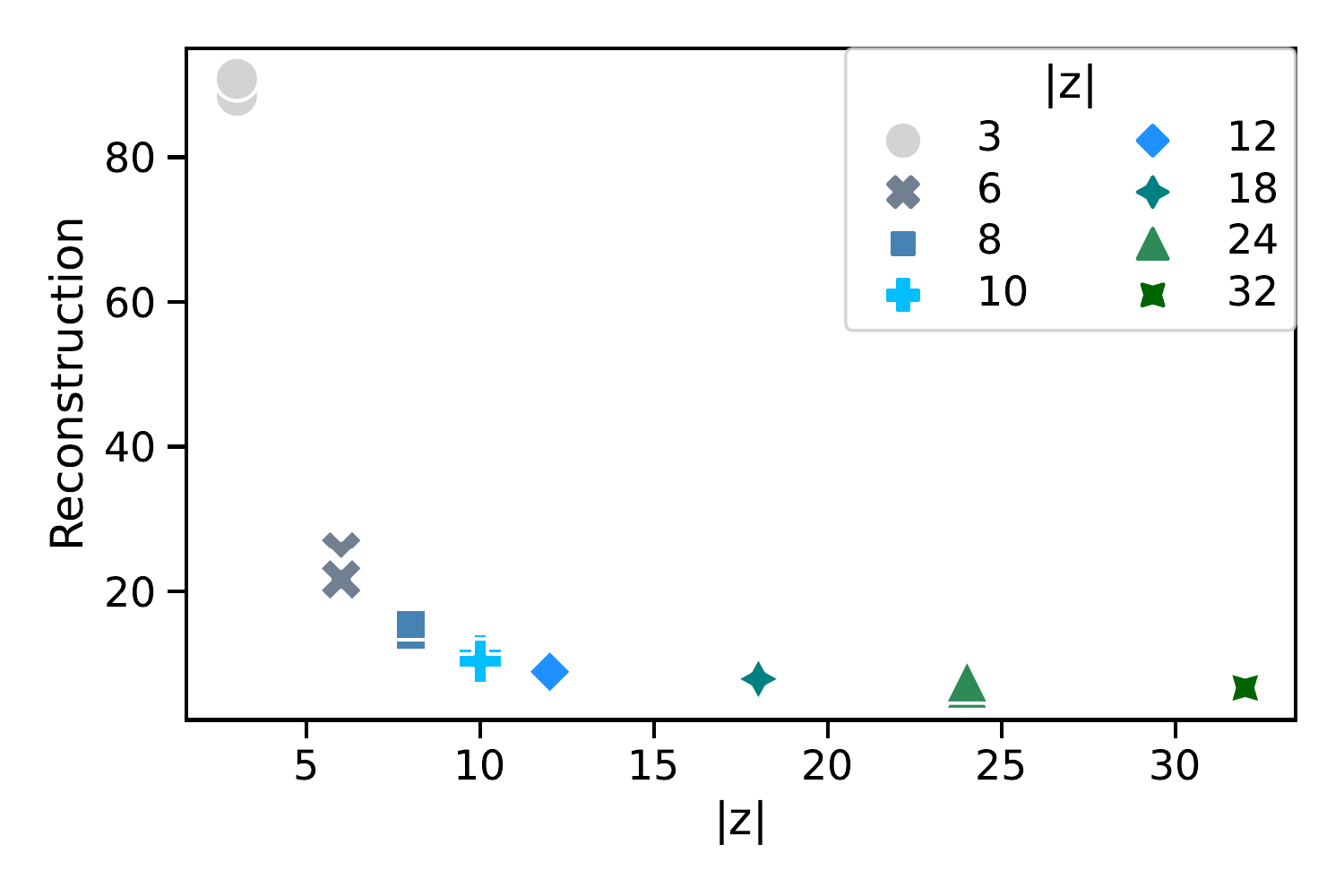}
        \caption{dSprites}
        \label{fig:dsprites-ae}
    \end{subfigure}%
    \begin{subfigure}{.33\textwidth}
        \centering
        \includegraphics[width=\linewidth]{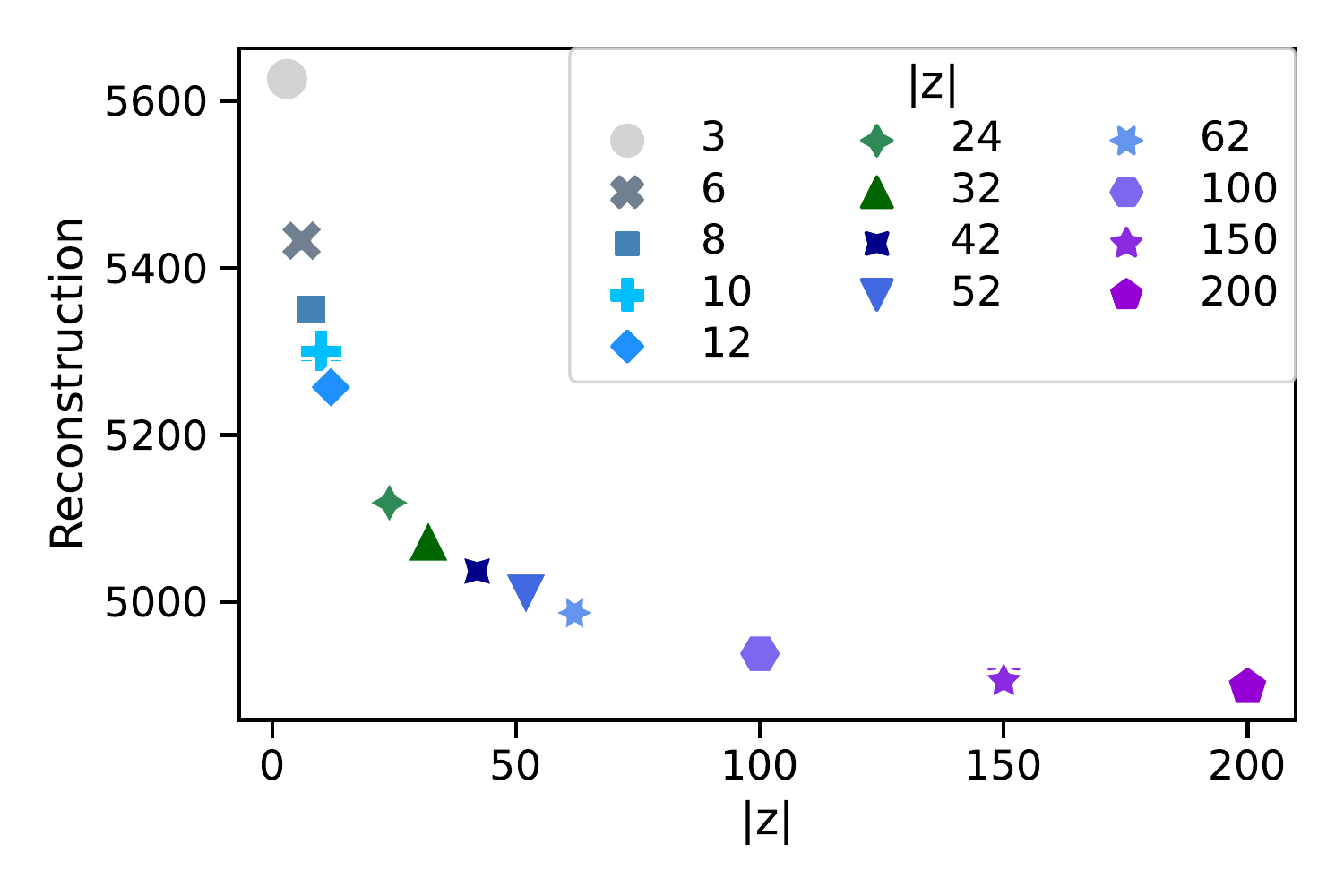}
        \caption{Celeba}
        \label{fig:celeba-ae}
    \end{subfigure}
    \caption{Reconstruction loss of AEs for Symsol, dSprites, and Celeba with an increasing number of latent variables.}
    \label{fig:fondue-rec-ae}
\end{figure}

\paragraph{Can FONDUE be applied to other architectures and learning objectives?}
We can see in~\Figref{fig:fondue-rec-ae} that deterministic AEs with equivalent architectures to the VAEs in~\Figref{fig:fondue-rec}
are performing well when provided with the same number of latent dimensions, indicating that FONDUE's results
could be reused for AEs trained on the same dataset with an identical architecture.
FONDUE also seems to be robust to architectural changes and worked equally well with fully-connected architectures (see~\Appref{sec:app-fc}).

    \section{Conclusion}\label{sec:conclusion}

By studying the ID estimates of the representations learned by VAEs, we have seen that the deeper the encoder's layers,
the lower their ID, while the ID of the decoder's layers consistently stayed close to the ID of the mean representation.
We also observed increasing discrepancies between mean and sampled IDs when the number of latent variable was large enough
for passive and mixed variables to appear.

This phenomenon is seen very early in the training process, and it leads to FONDUE:
an algorithm which can find the number of latent dimensions after which the mean and sampled representations
start to strongly diverge. After proving the correctness of our algorithm, we have shown that it is a computationally efficient alternative to grid search --- taking only
minutes to provide an estimation of the optimal number of dimensions to use --- which leads to a good tradeoff between the reconstruction
and regularisation losses. Moreover, FONDUE is not impacted by architectural changes, and its prediction can also be used
for deterministic autoencoders.
    \ifanonymous
    \else
        \subsubsection*{Acknowledgments}
The authors thank Th\'{e}ophile Champion and Declan Collins for their helpful comments on the paper.

    \fi

    \clearpage

    \bibliography{main}

\begin{thebibliography}{34}
\providecommand{\natexlab}[1]{#1}
\providecommand{\url}[1]{\texttt{#1}}
\expandafter\ifx\csname urlstyle\endcsname\relax
  \providecommand{\doi}[1]{doi: #1}\else
  \providecommand{\doi}{doi: \begingroup \urlstyle{rm}\Url}\fi

\bibitem[Ansuini et~al.(2019)Ansuini, Laio, Macke, and Zoccolan]{Ansuini2019}
Alessio Ansuini, Alessandro Laio, Jakob~H. Macke, and Davide Zoccolan.
\newblock {Intrinsic dimension of data representations in deep neural
  networks}.
\newblock In \emph{Advances in Neural Information Processing Systems},
  volume~32, 2019.

\bibitem[Arora et~al.(2018)Arora, Cohen, and Hazan]{Arora2018}
Sanjeev Arora, Nadav Cohen, and Elad Hazan.
\newblock On the optimization of deep networks: Implicit acceleration by
  overparameterization.
\newblock In Jennifer Dy and Andreas Krause (eds.), \emph{Proceedings of the
  35th International Conference on Machine Learning}, volume~80 of
  \emph{Proceedings of Machine Learning Research}, pp.\  244--253. PMLR, 2018.

\bibitem[Arvanitidis et~al.(2018)Arvanitidis, Hansen, and
  Hauberg]{Arvanitidis2018}
Georgios Arvanitidis, Lars~Kai Hansen, and S\o{}ren Hauberg.
\newblock {Latent Space Oddity: on the Curvature of Deep Generative Models}.
\newblock In \emph{International Conference on Learning Representations},
  volume~6, 2018.

\bibitem[Bonheme \& Grzes(2021)Bonheme and Grzes]{Bonheme2021}
Lisa Bonheme and Marek Grzes.
\newblock {Be More Active! Understanding the Differences between Mean and
  Sampled Representations of Variational Autoencoders}.
\newblock \emph{arXiv e-prints}, 2021.

\bibitem[Campadelli et~al.(2015)Campadelli, Casiraghi, Ceruti, and
  Rozza]{Campadelli2015}
P.~Campadelli, E.~Casiraghi, C.~Ceruti, and A.~Rozza.
\newblock Intrinsic dimension estimation: Relevant techniques and a benchmark
  framework.
\newblock \emph{Mathematical Problems in Engineering}, 2015, 2015.

\bibitem[Chollet(2021)]{Chollet2021}
F.~Chollet.
\newblock \emph{Deep Learning with Python, Second Edition}.
\newblock Manning, 2021.
\newblock ISBN 9781617296864.

\bibitem[Dai \& Wipf(2018)Dai and Wipf]{Dai2018}
Bin Dai and David Wipf.
\newblock {Diagnosing and Enhancing VAE Models}.
\newblock In \emph{International Conference on Learning Representations},
  volume~6, 2018.

\bibitem[Dai et~al.(2020)Dai, Wang, and Wipf]{Dai2020}
Bin Dai, Ziyu Wang, and David Wipf.
\newblock {The Usual Suspects? {R}eassessing Blame for {VAE} Posterior
  Collapse}.
\newblock In \emph{Proceedings of the 37th International Conference on Machine
  Learning}, 2020.

\bibitem[Doersch(2016)]{Doersch2016}
Carl Doersch.
\newblock {Tutorial on Variational Autoencoders}.
\newblock \emph{arXiv e-prints}, 2016.

\bibitem[Facco et~al.(2017)Facco, d\'Errico, Rodriguez, and Laio]{Facco2017}
Elena Facco, Maria d\'Errico, Alex Rodriguez, and Alessandro Laio.
\newblock Estimating the intrinsic dimension of datasets by a minimal
  neighborhood information.
\newblock \emph{Scientific Reports}, 7\penalty0 (1), 2017.

\bibitem[Falorsi et~al.(2018)Falorsi, De~Haan, Davidson, De~Cao, Weiler,
  Forr{\'e}, and Cohen]{Falorsi2018}
Luca Falorsi, Pim De~Haan, Tim~R Davidson, Nicola De~Cao, Maurice Weiler,
  Patrick Forr{\'e}, and Taco~S Cohen.
\newblock Explorations in homeomorphic variational auto-encoding.
\newblock \emph{arXiv e-prints}, 2018.

\bibitem[Gong et~al.(2019)Gong, Boddeti, and Jain]{Gong2019}
Sixue Gong, Vishnu~Naresh Boddeti, and Anil~K Jain.
\newblock On the intrinsic dimensionality of image representations.
\newblock In \emph{Proceedings of the IEEE/CVF Conference on Computer Vision
  and Pattern Recognition}, pp.\  3987--3996, 2019.

\bibitem[Hensel et~al.(2021)Hensel, Moor, and Rieck]{Hensel2021}
Felix Hensel, Michael Moor, and Bastian Rieck.
\newblock A survey of topological machine learning methods.
\newblock \emph{Frontiers in Artificial Intelligence}, 4, 2021.

\bibitem[Higgins et~al.(2017)Higgins, Matthey, Pal, Burgess, Glorot, Botvinick,
  Shakir, and Lerchner]{Higgins2017}
Irina Higgins, Loic Matthey, Arka Pal, Christopher Burgess, Xavier Glorot,
  Matthew Botvinick, Mohamed Shakir, and Alexander Lerchner.
\newblock {{$\beta$}-VAE: Learning Basic Visual Concepts with a Constrained
  Variational Framework}.
\newblock In \emph{International Conference on Learning Representations},
  volume~5, 2017.

\bibitem[Karbauskait{\.{e}} et~al.(2011)Karbauskait{\.{e}}, Dzemyda, and
  Maz{\.{e}}tis]{Karbauskaita2011}
Rasa Karbauskait{\.{e}}, Gintautas Dzemyda, and Edmundas Maz{\.{e}}tis.
\newblock Geodesic distances in the maximum likelihood estimator of intrinsic
  dimensionality.
\newblock \emph{Nonlinear Analysis}, 16\penalty0 (4):\penalty0 387--402, 2011.

\bibitem[Keller \& Welling(2021)Keller and Welling]{Keller2021}
T.~Anderson Keller and Max Welling.
\newblock Topographic {VAE}s learn equivariant capsules.
\newblock In A.~Beygelzimer, Y.~Dauphin, P.~Liang, and J.~Wortman Vaughan
  (eds.), \emph{Advances in Neural Information Processing Systems}, volume~35,
  2021.

\bibitem[Khrulkov \& Oseledets(2018)Khrulkov and Oseledets]{Khrulkov2018}
Valentin Khrulkov and Ivan Oseledets.
\newblock Geometry score: A method for comparing generative adversarial
  networks.
\newblock In Jennifer Dy and Andreas Krause (eds.), \emph{Proceedings of the
  35th International Conference on Machine Learning}, volume~80 of
  \emph{Proceedings of Machine Learning Research}, pp.\  2621--2629. PMLR,
  10--15 Jul 2018.

\bibitem[Kingma \& Welling(2014)Kingma and Welling]{Kingma2013}
Diederik~P. Kingma and Max Welling.
\newblock {Auto-Encoding Variational Bayes}.
\newblock In \emph{International Conference on Learning Representations},
  volume~2, 2014.

\bibitem[Levina \& Bickel(2004)Levina and Bickel]{Levina2004}
Elizaveta Levina and Peter~J. Bickel.
\newblock {Maximum Likelihood Estimation of Intrinsic Dimension}.
\newblock In \emph{Advances in Neural Information Processing Systems},
  volume~16, 2004.

\bibitem[Liu et~al.(2015)Liu, Luo, Wang, and Tang]{Liu2015}
Ziwei Liu, Ping Luo, Xiaogang Wang, and Xiaoou Tang.
\newblock Deep learning face attributes in the wild.
\newblock In \emph{ICCV}, 2015.

\bibitem[Locatello et~al.(2019)Locatello, Bauer, Lucic, Raetsch, Gelly,
  Sch{\"{o}}lkopf, and Bachem]{Locatello2019a}
Francesco Locatello, Stefan Bauer, Mario Lucic, Gunnar Raetsch, Sylvain Gelly,
  Bernhard Sch{\"{o}}lkopf, and Olivier Bachem.
\newblock {Challenging Common Assumptions in the Unsupervised Learning of
  Disentangled Representations}.
\newblock In \emph{Proceedings of the 36th International Conference on Machine
  Learning}, volume~97 of \emph{Proceedings of Machine Learning Research},
  2019.

\bibitem[Lucas et~al.(2019{\natexlab{a}})Lucas, Tucker, Grosse, and
  Norouzi]{Lucas2019}
James Lucas, George Tucker, Roger~B. Grosse, and Mohammad Norouzi.
\newblock {Understanding Posterior Collapse in Generative Latent Variable
  Models}.
\newblock In \emph{Deep Generative Models for Highly Structured Data, {ICLR}
  2019 Workshop}, 2019{\natexlab{a}}.

\bibitem[Lucas et~al.(2019{\natexlab{b}})Lucas, Tucker, Grosse, and
  Norouzi]{Lucas2019b}
James Lucas, George Tucker, Roger~B. Grosse, and Mohammad Norouzi.
\newblock {Don't Blame the ELBO! A linear VAE Perspective on Posterior
  Collapse}.
\newblock In \emph{Advances in Neural Information Processing Systems},
  volume~32, 2019{\natexlab{b}}.

\bibitem[MacKay \& Ghahramani(2005)MacKay and Ghahramani]{Mackay2005}
David~J.C. MacKay and Zoubin Ghahramani.
\newblock {Comments on ‘Maximum Likelihood Estimation of Intrinsic
  Dimension’ by E. Levina and P. Bickel (2004)}, 2005.
\newblock URL \url{http://www.inference.org.uk/mackay/dimension/}.

\bibitem[Maheswaranathan et~al.(2019)Maheswaranathan, Williams, Golub, Ganguli,
  and Sussillo]{Maheswaranathan2019}
Niru Maheswaranathan, Alex Williams, Matthew Golub, Surya Ganguli, and David
  Sussillo.
\newblock {Universality and Individuality in Neural Dynamics Across Large
  Populations of Recurrent Networks}.
\newblock In \emph{Advances in Neural Information Processing Systems},
  volume~32, 2019.

\bibitem[Murphy et~al.(2021)Murphy, Esteves, Jampani, Ramalingam, and
  Makadia]{Murphy2021}
Kieran Murphy, Carlos Esteves, Varun Jampani, Srikumar Ramalingam, and Ameesh
  Makadia.
\newblock Implicit representation of probability distributions on the rotation
  manifold.
\newblock In \emph{International Conference on Machine Learning}, 2021.

\bibitem[Naitzat et~al.(2020)Naitzat, Zhitnikov, and Lim]{Naitzat2020}
Gregory Naitzat, Andrey Zhitnikov, and Lek-Heng Lim.
\newblock {Topology of Deep Neural Networks}.
\newblock \emph{Journal of Machine Learning Research}, 21\penalty0 (184), 2020.

\bibitem[{Perez Rey} et~al.(2020){Perez Rey}, Menkovski, and
  Portegies]{PerezRey2020}
{Luis A.} {Perez Rey}, V.~Menkovski, and {Jacobus W.} Portegies.
\newblock {Diffusion Variational Autoencoders}.
\newblock In \emph{29th International Joint Conference on Artificial
  Intelligence - 17th Pacific Rim International Conference on Artificial
  Intelligence. (IJCAI-PRICAI 2020)}, 2020.

\bibitem[Pope et~al.(2021)Pope, Zhu, Abdelkader, Goldblum, and
  Goldstein]{Pope2021}
Phillip Pope, Chen Zhu, Ahmed Abdelkader, Micah Goldblum, and Tom Goldstein.
\newblock {The Intrinsic Dimension of Images and Its Impact on Learning}.
\newblock In \emph{International Conference on Learning Representations},
  volume~9, 2021.

\bibitem[Rezende \& Mohamed(2015)Rezende and Mohamed]{Rezende2015}
Danilo Rezende and Shakir Mohamed.
\newblock {Variational Inference with Normalizing Flows}.
\newblock In \emph{Proceedings of the 32nd International Conference on Machine
  Learning}, volume~37 of \emph{Proceedings of Machine Learning Research},
  2015.

\bibitem[Rieck et~al.(2019)Rieck, Togninalli, Bock, Moor, Horn, Gumbsch, and
  Borgwardt]{Rieck2018}
Bastian Rieck, Matteo Togninalli, Christian Bock, Michael Moor, Max Horn,
  Thomas Gumbsch, and Karsten Borgwardt.
\newblock {Neural Persistence: A Complexity Measure for Deep Neural Networks
  Using Algebraic Topology}.
\newblock In \emph{International Conference on Learning Representations},
  volume~7, 2019.

\bibitem[Rolinek et~al.(2019)Rolinek, Zietlow, and Martius]{Rolinek2019}
Michal Rolinek, Dominik Zietlow, and Georg Martius.
\newblock {Variational Autoencoders Pursue {PCA} Directions (by Accident)}.
\newblock In \emph{Proceedings of the IEEE/CVF Conference on Computer Vision
  and Pattern Recognition (CVPR)}, 2019.

\bibitem[Sankararaman et~al.(2020)Sankararaman, De, Xu, Huang, and
  Goldstein]{Sankararaman2020}
Karthik~Abinav Sankararaman, Soham De, Zheng Xu, W.~Ronny Huang, and Tom
  Goldstein.
\newblock The impact of neural network overparameterization on gradient
  confusion and stochastic gradient descent.
\newblock In Hal~Daumé III and Aarti Singh (eds.), \emph{Proceedings of the
  37th International Conference on Machine Learning}, volume 119 of
  \emph{Proceedings of Machine Learning Research}, pp.\  8469--8479. PMLR,
  2020.

\bibitem[Zhou et~al.(2021)Zhou, Zelikman, Lu, Ng, Carlsson, and
  Ermon]{Zhou2021}
Sharon Zhou, Eric Zelikman, Fred Lu, Andrew~Y. Ng, Gunnar~E. Carlsson, and
  Stefano Ermon.
\newblock Evaluating the disentanglement of deep generative models through
  manifold topology.
\newblock In \emph{International Conference on Learning Representations},
  volume~9, 2021.

\end{thebibliography}

    \clearpage
    \appendix
    \section{Proof of~\Thref{thm:fondue}}\label{sec:app-fondue-proof}

This section provides the full proof of~\Thref{thm:fondue}. To ease its reading, let us first define an axiom
based on our observation from~\Secref{subsec:res-vaes} that the IDEs of the mean and sampled representations start
to diverge only after the number of latent dimensions has become large enough for (unused) passive variables to appear.

\begin{axm}\label{axm:fondue}
    Let $IDE_x^y$ be the IDE of layer $x$ using $y$ latent dimensions. Given the sets $\sA = \{a | IDE^{a}_z - IDE^{a}_{\mu} \leqslant threshold\}$ and $\sB = \{b | IDE^{b}_z - IDE^{b}_{\mu} > threshold\}$,
    we have $a < b \quad  \forall\, a \in \sA, \; \forall\, b \in \sB$.
\end{axm}

\begin{rem}\label{rem:fondue}
    Given that $l$ and $u$ only take values of latent dimensions for which $IDE_z - IDE_{\mu} \leqslant threshold$ and $IDE_z - IDE_{\mu} > threshold$, respectively,
    \Axref{axm:fondue} implies that for all iterations $i$, $l_i \in \sA$ and $u_i \in \sB$ and $l_i < u_i$.
\end{rem}

Using the loop invariant $l_i \leqslant p_i \leqslant u_i$ for each iteration $i$, we will now show that~\Algref{alg:fondue}
terminates when $l_i = p_i = \operatorname{floor}\left(\frac{l_i + u_i}{2}\right)$, which can only be reached when $u_i = p_i + 1$,
that is when $p_i$ is the maximum number of latent dimensions for which we have $IDE_z - IDE_{\mu} \leqslant threshold$.

\begin{proof}
    $ $\\\\
    \textbf{Initialisation:} $l_0 = 0, p_0 = IDE_{data}, u_0 = \infty$, thus $l_0 < p_0 < u_0$.\\\\
    \textbf{Maintenance:} We will consider both branches of the if statement separately:
    \begin{itemize}
        \item For $IDE_z - IDE_{\mu} \leqslant threshold$ (lines 9-11), $u_i = u_{i-1}$, $p_i = min(p_{i-1} \times 2, u_i)$, and $l_i = p_{i-1}$.
        We directly see that $p_i \leqslant u_i$. We know from~\Remref{rem:fondue} that $l_i < u_i$ and we also
        have $l_i < p_{i-1} \times 2$, it follows that $l_i < p_i$. Grouping both inequalities, we get $l_i < p_i \leqslant u_i$.
        \item For $IDE_z - IDE_{\mu} > threshold$ (lines 12-14), $u_i = p_{i-1}, l_i=l_{i-1}$, and $p_i = \operatorname{floor}\left(\frac{l_i + u_i}{2}\right)$.
        Using~\Remref{rem:fondue} we can directly see that $l_i \leqslant \operatorname{floor}\left(\frac{l_i + u_i}{2}\right) < u_i$ and
        we obtain $l_i \leqslant p_i < u_i$.
    \end{itemize}
    \textbf{Termination:} The loop terminates when $l_i = p_i$.
    Given that $l_i < p_i$ when $IDE_z - IDE_{\mu} \leqslant threshold$, this is only possible when $IDE_z - IDE_{\mu} > threshold$, which is when $p_i = \operatorname{floor}\left(\frac{l_i + u_i}{2}\right)$.
    We know from~\Remref{rem:fondue} that $l_i < u_i$, so we must have $(l_i + u_i) \operatorname{mod} 2 > 0$.
    As $a \operatorname{mod} 2 \in \{0, 1\}$, the only possible value for $u_i$ to satisfy $l_i = p_i = \operatorname{floor}\left(\frac{l_i + u_i}{2}\right)$ is $u_i = p_i + 1$.
    Thus, $p_i$ is the largest number of latent dimensions for which $IDE_z - IDE_{\mu} \leqslant threshold$.

\end{proof}
    \section{Resources}\label{sec:app-resources}
As mentioned in~\Twosecrefs{sec:intro}{sec:xp}, we released the code of our experiment, and the IDEs:
\ifanonymous
\else
\begin{itemize}
    \item The IDEs can be downloaded at~\url{https://data.kent.ac.uk/id/eprint/455}
    \item \texttt{symsol\_reduced}, the reduced version of Symmetric solids can be downloaded at~\url{https://data.kent.ac.uk/436}
    \item The code is available at~\url{https://github.com/bonheml/VAE_learning_dynamics}
\end{itemize}
\fi

    \section{Experimental setup}\label{sec:xp-setup}

Our implementation uses the same hyperparameters as~\citet{Locatello2019a}, as listed in~\Tableref{table:global-hyperparam}.
We reimplemented the~\citet{Locatello2019a} code base, designed for Tensorflow 1, in Tensorflow 2 using Keras.
The model architectures used are also similar, as described in~\Twotablerefs{table:architecture}{table:fc-architecture}.
We used the convolutional architecture in the main paper and the fully-connected architecture in~\Appref{sec:app-fc}.
Each model is trained 5 times with seed values from 0 to 4.
Every image input is normalised to have pixel values between 0 and 1.
TwoNN is used with an anchor of 0.9, and the hyperparameters for MLE can be found in~\Tableref{table:mle-hyperparam}.

\begin{table}[ht!]
    \centering
    \caption{VAEs hyperparameters}
    \label{table:global-hyperparam}
    \begin{tabular}{ l l }
        \hline
        Parameter & Value \\
        \hline
        \rule{0pt}{2.6ex}Batch size & 64  \\
        Latent space dimension & 3, 6, 8, 10, 12, 18, 24, 32. \\&For Celeba only: 42, 52, 62, 100, 150, 200  \\
        Optimizer & Adam \\
        Adam: $\beta_1$ & 0.9 \\
        Adam: $\beta_2$ & 0.999 \\
        Adam: $\epsilon$ & 1e-8 \\
        Adam: learning rate & 0.0001 \\
        Reconstruction loss & Bernoulli \\
        Training steps & 300,000 \\
        Train/test split & 90/10\\
        $\beta$ & 1\\
        \hline
    \end{tabular}
\end{table}

\begin{table}[ht!]
    \centering
    \caption{Architecture}
    \label{table:architecture}
    \begin{tabularx}{\linewidth}{ X X }
        \hline
        Encoder & Decoder \\
        \hline
        \rule{0pt}{2.6ex}Input: $\R^{64 \times 63 \times channels}$ & $\R^{10}$ \\
        Conv, kernel=4×4, filters=32, activation=ReLU, strides=2 & FC, output shape=256, activation=ReLU \\
        Conv, kernel=4×4, filters=32, activation=ReLU, strides=2 & FC, output shape=4x4x64, activation=ReLU \\
        Conv, kernel=4×4, filters=64, activation=ReLU, strides=2 & Deconv, kernel=4×4, filters=64, activation=ReLU, strides=2 \\
        Conv, kernel=4×4, filters=64, activation=ReLU, strides=2 & Deconv, kernel=4×4, filters=32, activation=ReLU, strides=2 \\
        FC, output shape=256, activation=ReLU, strides=2 & Deconv, kernel=4×4, filters=32, activation=ReLU, strides=2 \\
        FC, output shape=2x10 & Deconv, kernel=4×4, filters=channels, activation=ReLU, strides=2 \\
        \hline
    \end{tabularx}
\end{table}

\begin{table}[ht!]
    \centering
    \caption{Fully-connected architecture}
    \label{table:fc-architecture}
    \begin{tabularx}{\linewidth}{ X X }
        \hline
        Encoder & Decoder \\
        \hline
        \rule{0pt}{2.6ex}Input: $\R^{64 \times 63 \times channels}$ & $\R^{10}$ \\
        FC, output shape=1200, activation=ReLU & FC, output shape=256, activation=tanh \\
        FC, output shape=1200, activation=ReLU & FC, output shape=1200, activation=tanh \\
        FC, output shape=2x10 & FC, output shape=1200, activation=tanh \\
        \hline
    \end{tabularx}
\end{table}

\begin{table}[ht!]
    \centering
    \caption{MLE hyperparameters}
    \label{table:mle-hyperparam}
    \begin{tabular}{ l l }
        \hline
        Parameter & Value \\
        \hline
        \rule{0pt}{2.6ex}k & 3, 5, 10, 20\\
        anchor & 0.8\\
        seed & 0\\
        runs & 5\\
        \hline
    \end{tabular}
\end{table}
    \section{FONDUE on fully-connected architectures}\label{sec:app-fc}

\begin{figure}[ht!]
    \centering
    \begin{subfigure}{.33\textwidth}
        \centering
        \includegraphics[width=\linewidth]{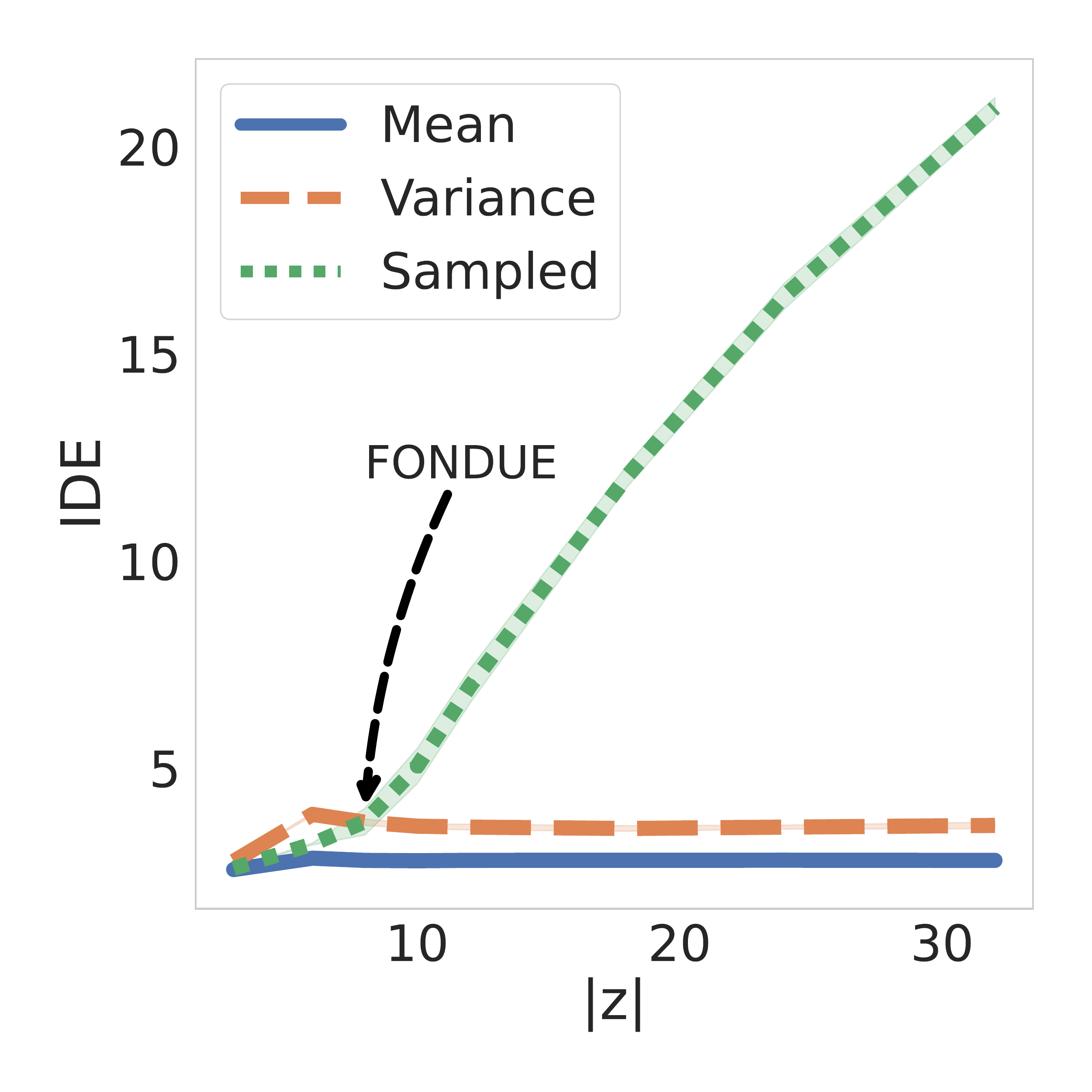}
        \caption{Symsol}
        \label{fig:fondue-ide-symsol-fc}
    \end{subfigure}%
    \begin{subfigure}{.33\textwidth}
        \centering
        \includegraphics[width=\linewidth]{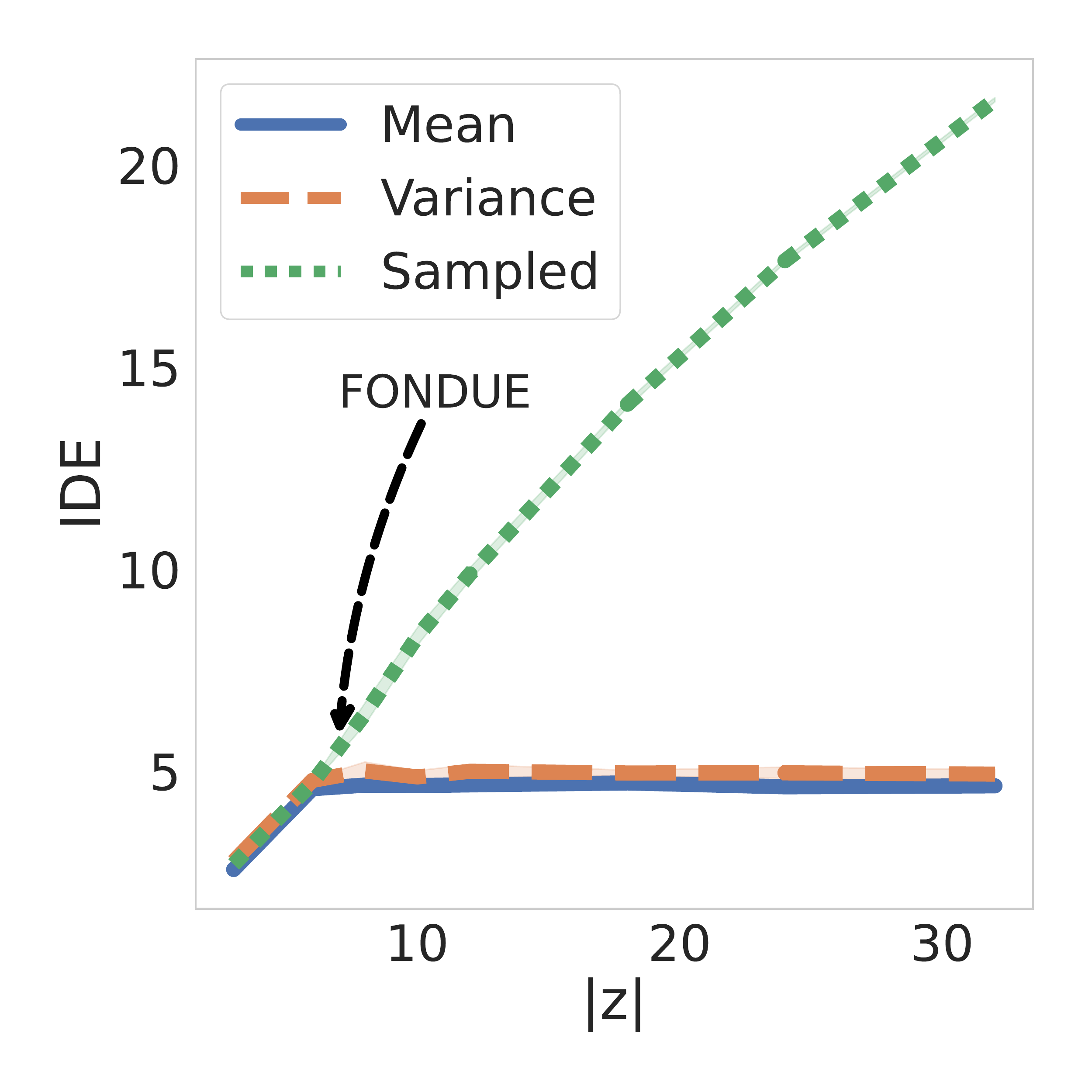}
        \caption{dSprites}
        \label{fig:fondue-ide-dsprites-fc}
    \end{subfigure}%
    \begin{subfigure}{.33\textwidth}
        \centering
        \includegraphics[width=\linewidth]{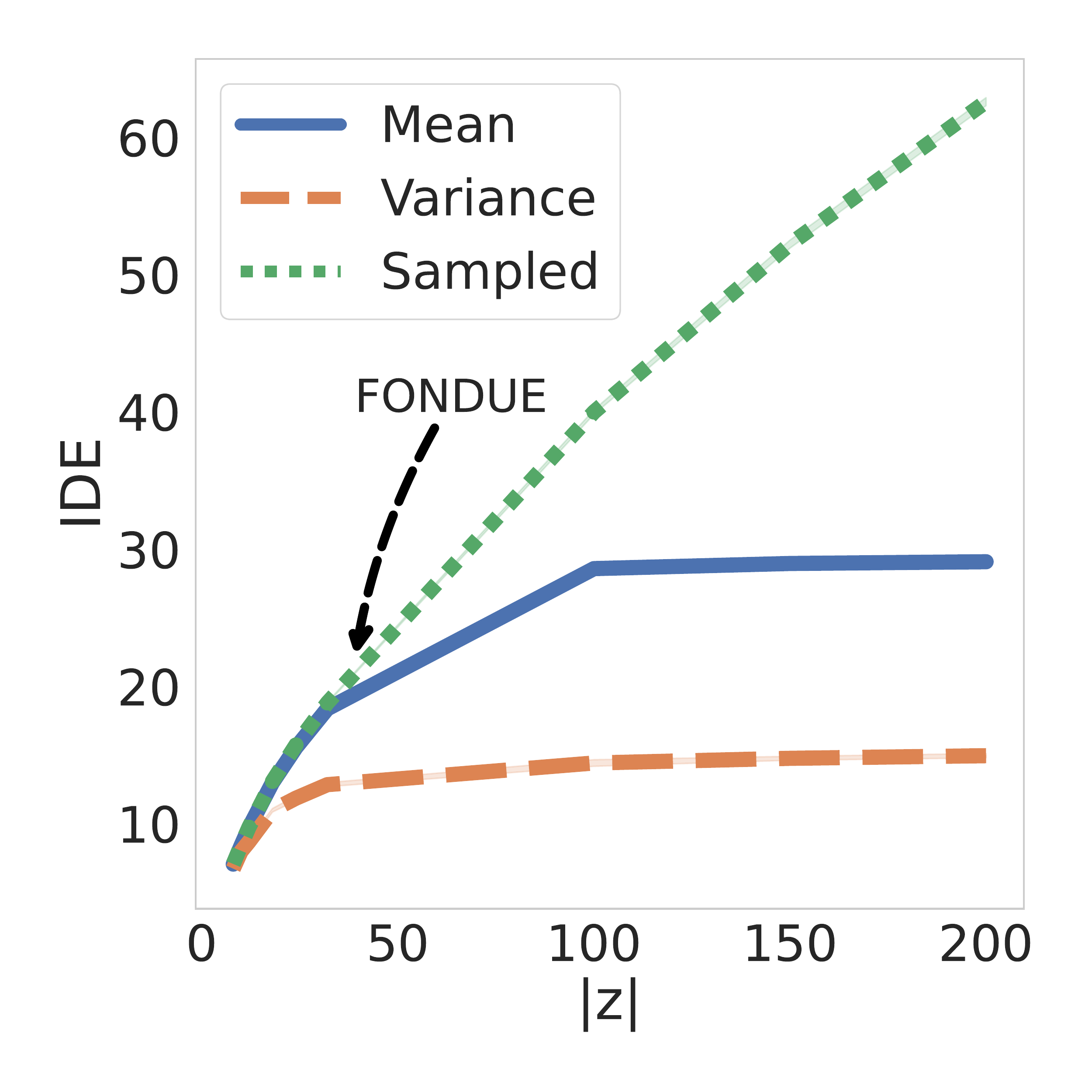}
        \caption{Celeba}
        \label{fig:fondue-ide-celeba-fc}
    \end{subfigure}
    \caption{Number of latent dimensions provided by FONDUE for fully-connected VAEs: $|\rvz| = 8$ on Symsol, $|\rvz| = 7$ on dSprites, and $|\rvz| = 39$ on Celeba.}
    \label{fig:fondue-ide-fc}
\end{figure}

We report the results obtained by FONDUE for fully-connected (FC) architectures in~\Tableref{table:fondue-ides-fc}
and~\Figref{fig:fondue-ide-fc}.
As shown in~\Tableref{table:fondue-ides-fc}, the execution time for finding the optimal number of dimensions of a
dataset is much shorter than for training one model (this is approximately 2h on the same GPUs), in similarity with convolutional VAEs.
As in~\Secref{subsec:res-fondue}, FONDUE correctly finds the number of latent dimensions after which the mean and sampled
IDEs start to diverge, as shown in~\Figref{fig:fondue-ide-fc}.
One can see that the number of latent variables needed for FC VAEs is much lower than for convolutional VAEs (see~\Tableref{table:fondue-ides} for comparison).
For dSprites, it is near the true ID of the data, and for Celeba, it is close to the data IDE reported in~\Figref{fig:ide-data} of \Secref{subsec:res-data}.

As in~\Secref{subsec:res-fondue}, we gradually increase the number of epochs until FONDUE reaches a stable estimation
of the latent dimensions.
As these models have fewer parameters than the convolutional architecture used in~\Secref{subsec:res-fondue}, they converge more slowly and need
to be trained for more epochs on Celeba and Symsol before reaching a stable estimation~\citep{Arora2018,Sankararaman2020}.
dSprites contains more data examples than the other datasets and less complex data than Celeba, which can explain its quicker convergence.

\begin{figure}[ht!]
    \centering
    \begin{subfigure}{.33\textwidth}
        \centering
        \includegraphics[width=\linewidth]{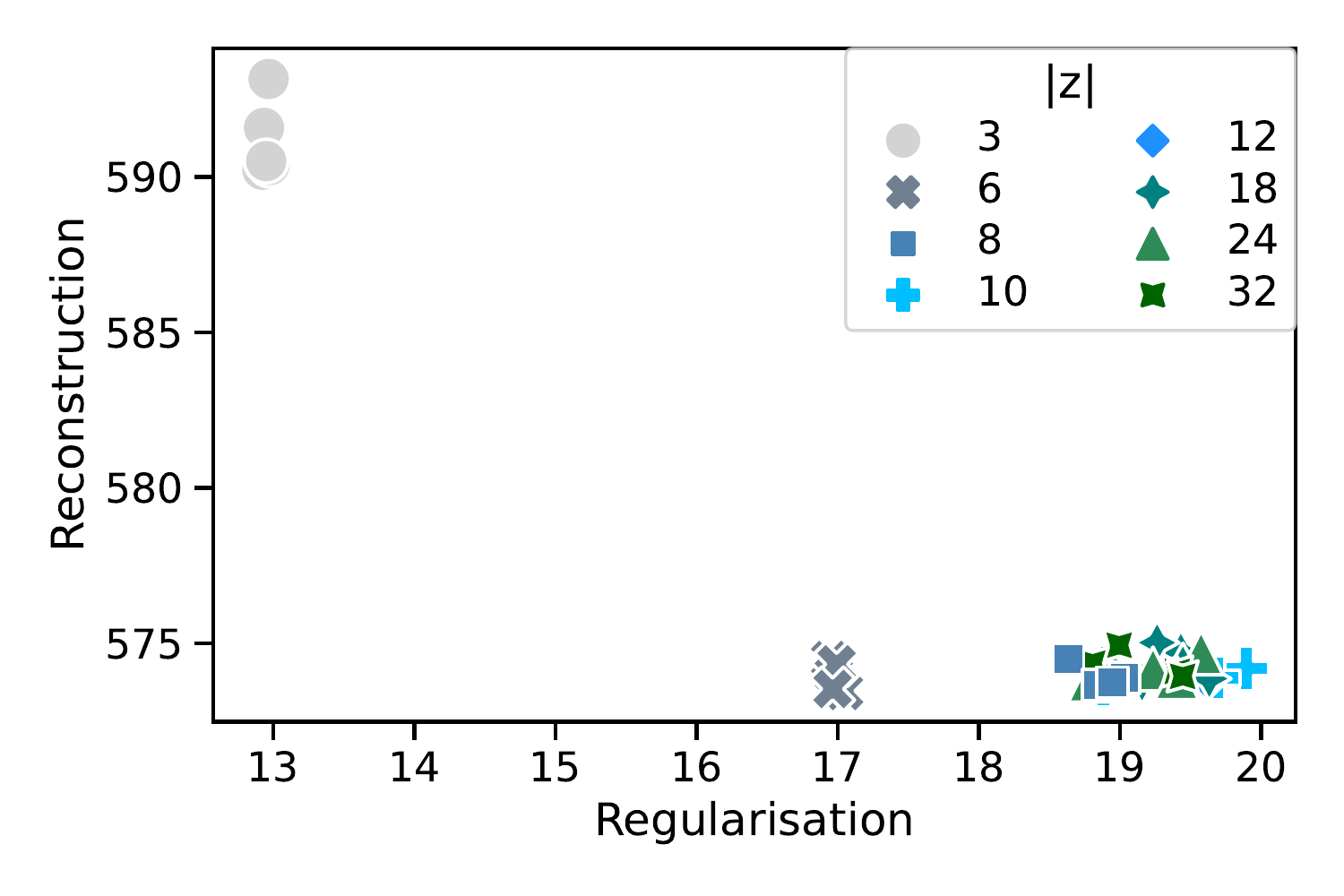}
        \caption{Symsol}
        \label{fig:fondue-rec-symsol-fc}
    \end{subfigure}%
    \begin{subfigure}{.33\textwidth}
        \centering
        \includegraphics[width=\linewidth]{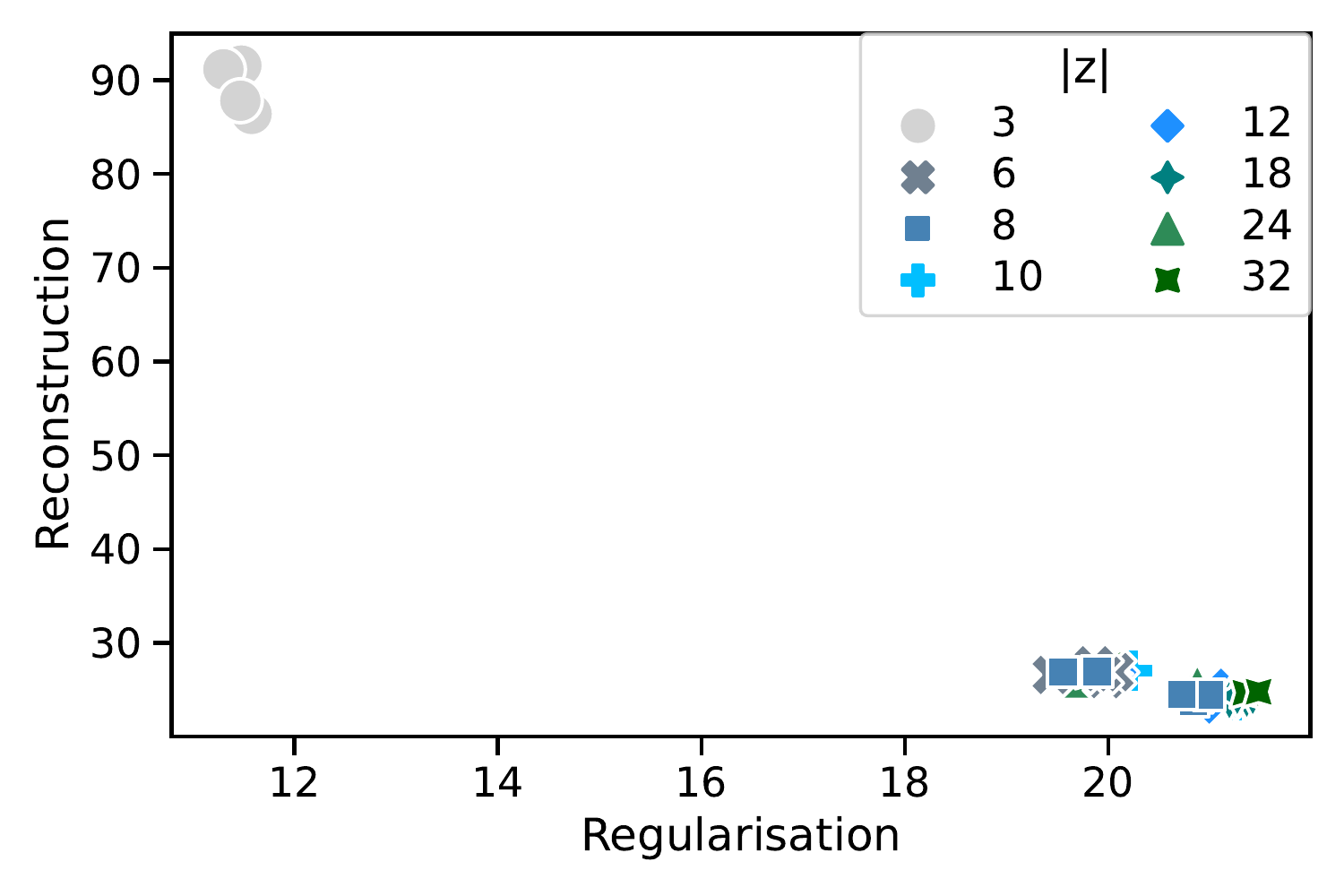}
        \caption{dSprites}
        \label{fig:fondue-rec-dsprites-fc}
    \end{subfigure}%
    \begin{subfigure}{.33\textwidth}
        \centering
        \includegraphics[width=\linewidth]{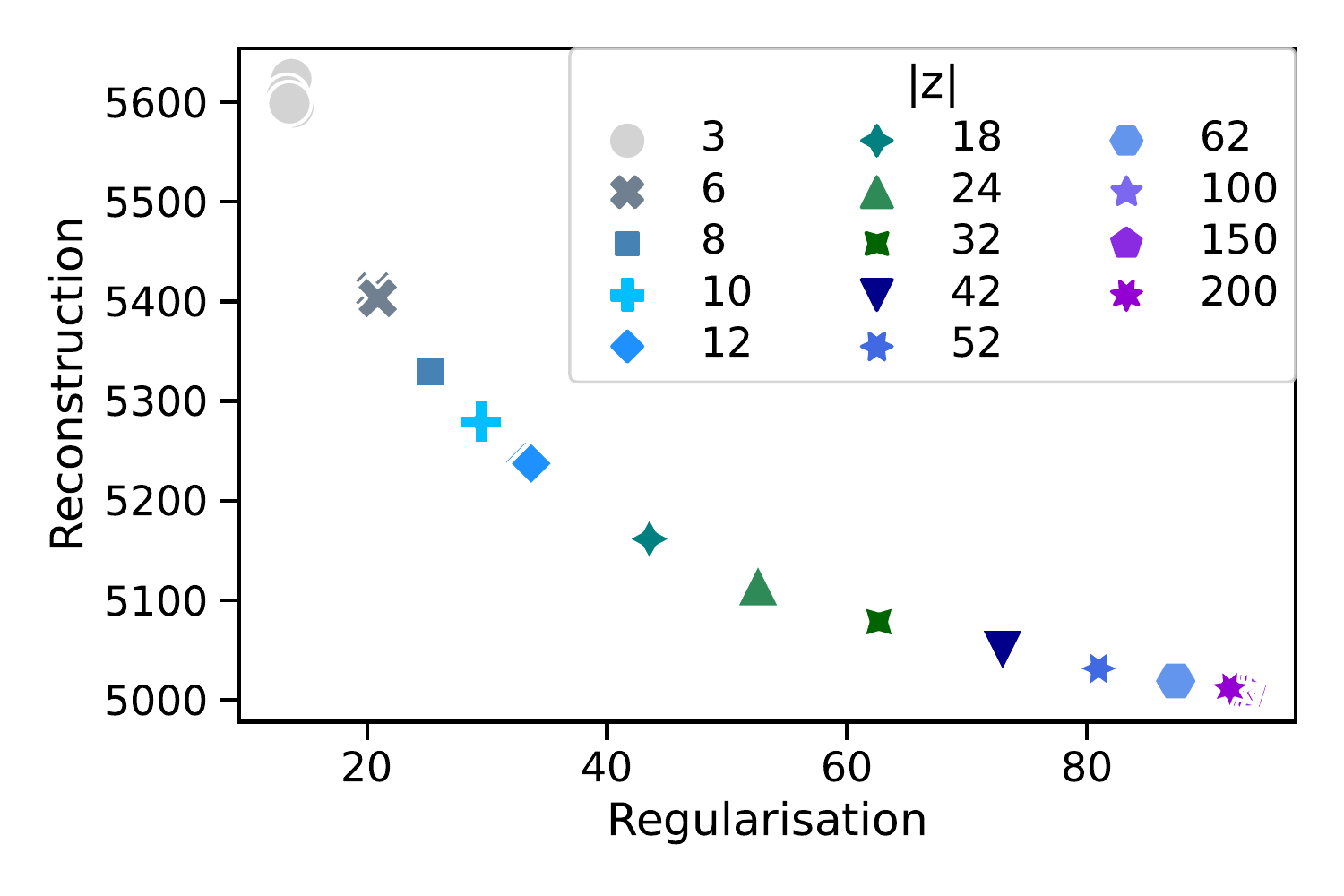}
        \caption{Celeba}
        \label{fig:fondue-rec-celeba-fc}
    \end{subfigure}
    \caption{Reconstruction and regularisation scores of fully-connected VAEs for Symsol, dSprites, and Celeba with an increasing number of latent variables $|\rvz|$.}
    \label{fig:fondue-rec-fc}
\end{figure}

\begin{figure}[ht!]
    \centering
    \begin{subfigure}{.33\textwidth}
        \centering
        \includegraphics[width=\linewidth]{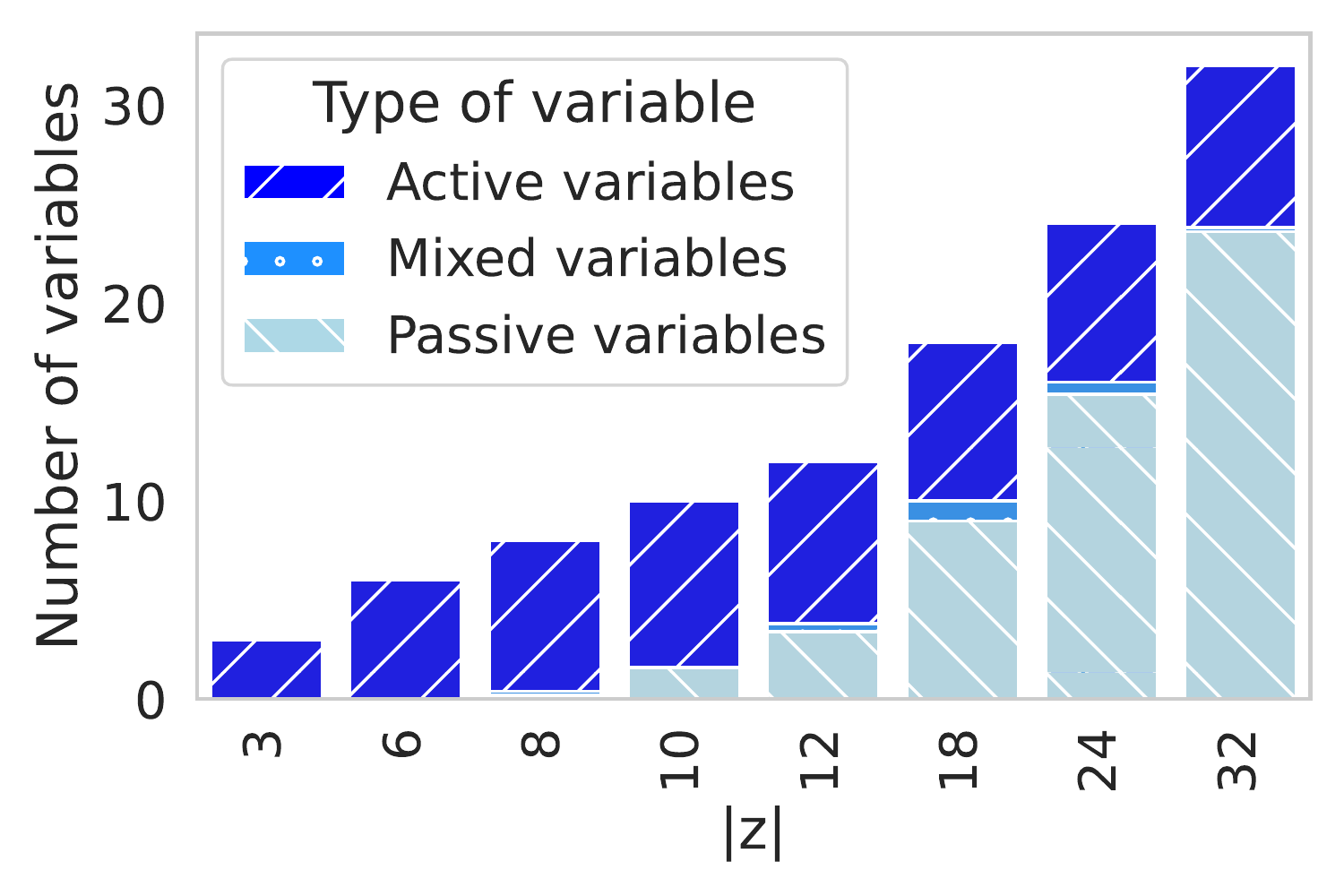}
        \caption{Symsol}
        \label{fig:var-type-symsol-fc}
    \end{subfigure}%
    \begin{subfigure}{.33\textwidth}
        \centering
        \includegraphics[width=\linewidth]{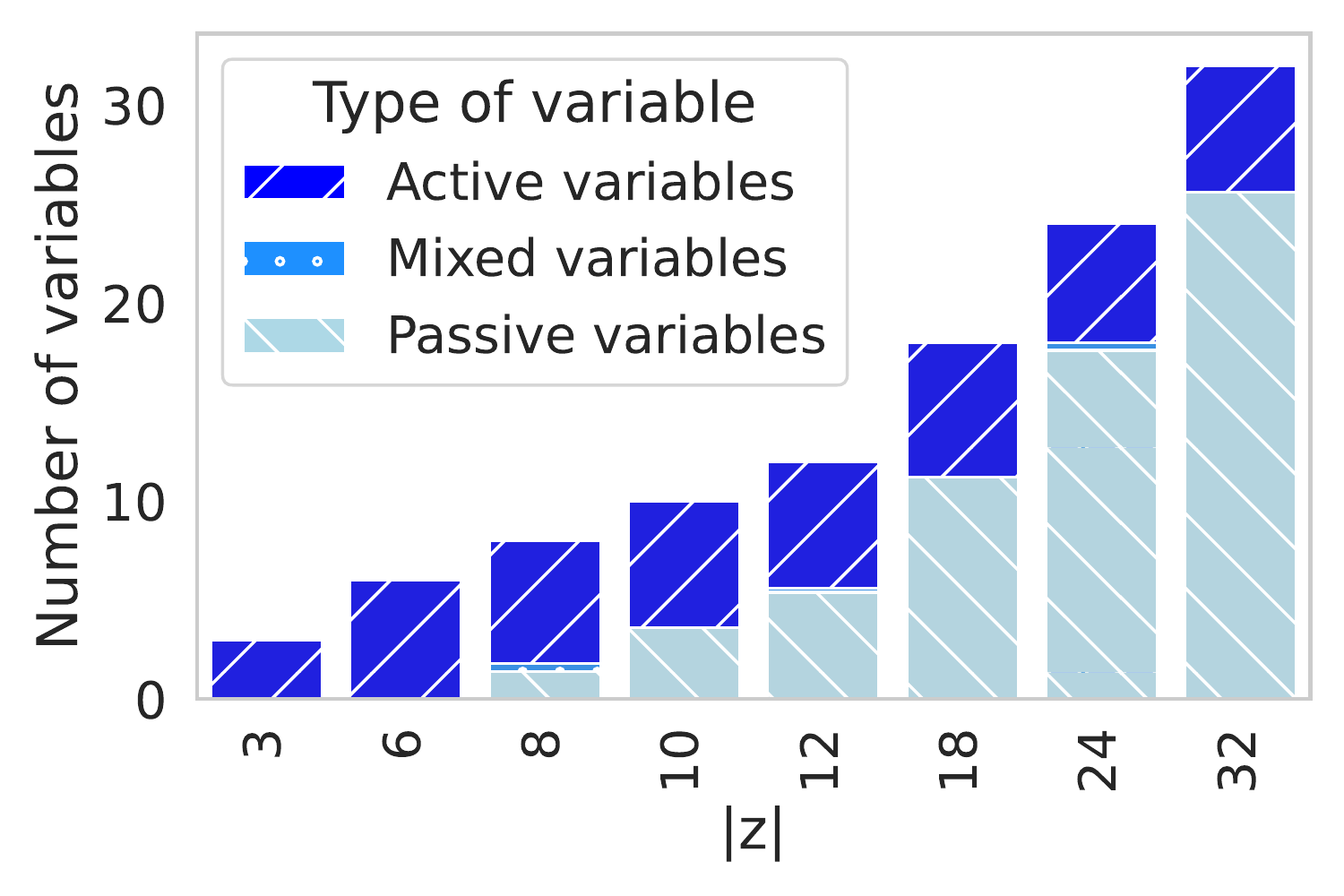}
        \caption{dSprites}
        \label{fig:var-type-dsprites-fc}
    \end{subfigure}%
    \begin{subfigure}{.33\textwidth}
        \centering
        \includegraphics[width=\linewidth]{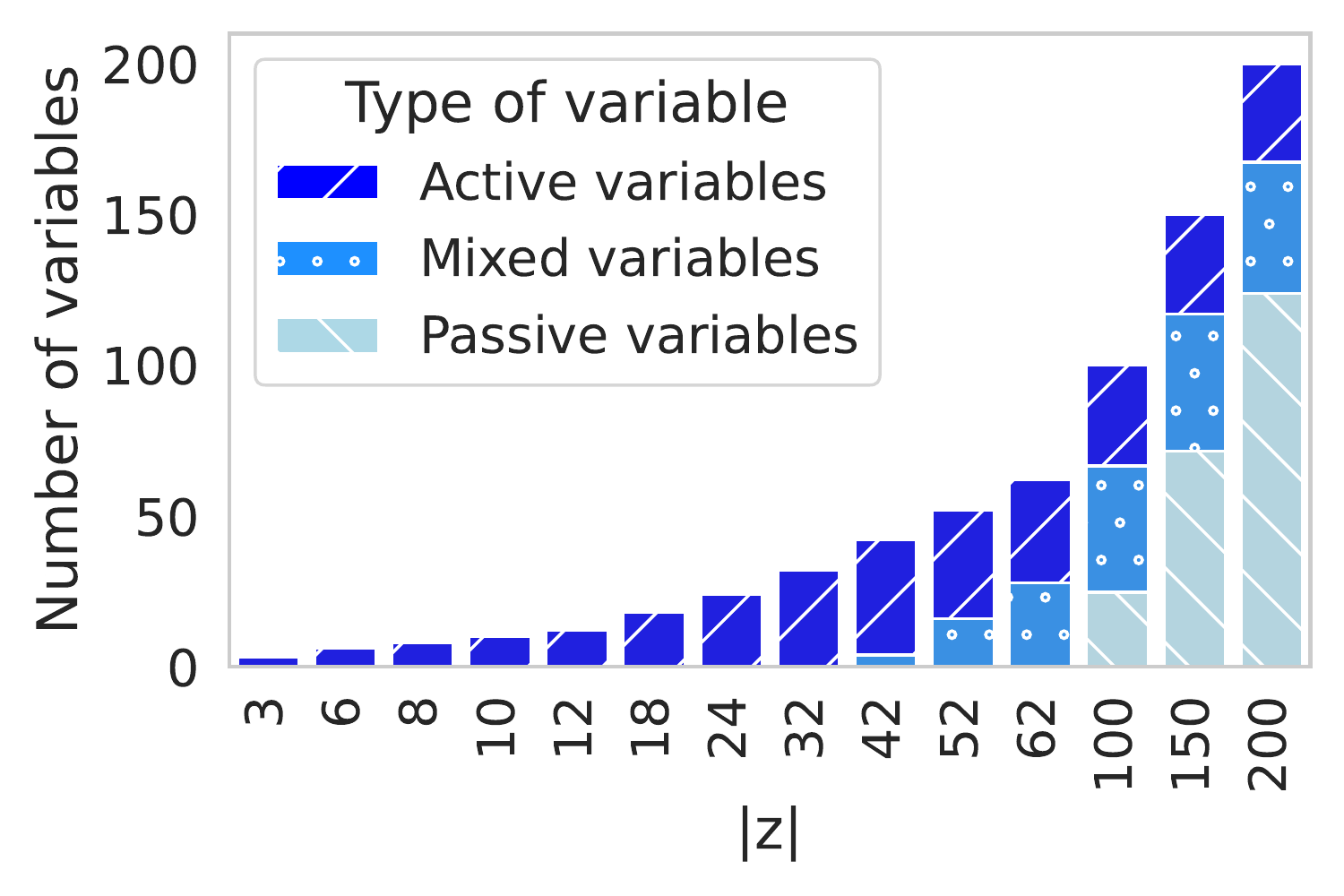}
        \caption{Celeba}
        \label{fig:var-type-celeba-fc}
    \end{subfigure}
    \caption{Quantity of active, mixed, and passive variables of VAEs trained with an increasing number of latent dimensions $|\rvz|$.
    (a), (b), and (c) show the results on Symsol,
        dSprites, and Celeba, respectively.}
    \label{fig:var-type-fc}
\end{figure}

For dSprites and Symsol, the number of dimensions selected by FONDUE corresponds to the number of dimensions after which the
reconstruction stops improving and the regularisation loss remains stable (see~\Figref{fig:fondue-rec-fc}).
For Celeba, the reconstruction continues to improve slightly after 39 latent dimensions, due to the addition of
variables between 42 and 100 latent dimensions, as illustrated in~\Figref{fig:var-type-fc}.
As in convolutional architectures, one could increase the threshold of FONDUE to take more mixed variables into account.

Overall, we can see that FONDUE also provides good results on the FC architectures, despite a slower convergence, showing robustness to architectural changes.

\begin{table}[ht!]
    \centering
    \caption{Number of latent variables obtained with FONDUE for fully-connected architectures.
    The results are averaged over 5 seeds, and computation times are reported for NVIDIA A100 GPUs.
    The computation time is given for one run of FONDUE over the minimum number of epochs needed to obtain a stable score.}
    \label{table:fondue-ides-fc}
    \begin{tabular}{ l l l l l}
        \hline
        Dataset & Dimensionality (avg $\pm$ SD) & Time/run & Models trained & Epochs/training\\
        \hline
        \rule{0pt}{2.6ex}Symsol & 8 $\pm$ 0 &  15 min & 6 & 6\\
        dSprites & 6.6 $\pm$ 0.5 & 16 min & 5 & 1\\
        Celeba &  39 $\pm$ 0.6 & 50 min & 7 & 9\\
        \hline
    \end{tabular}
\end{table}

    \section{Additional details on mean, variance, and sampled representations}\label{sec:app-vae}
\begin{figure}[ht!]
    \centering
    \includegraphics[width=0.7\textwidth]{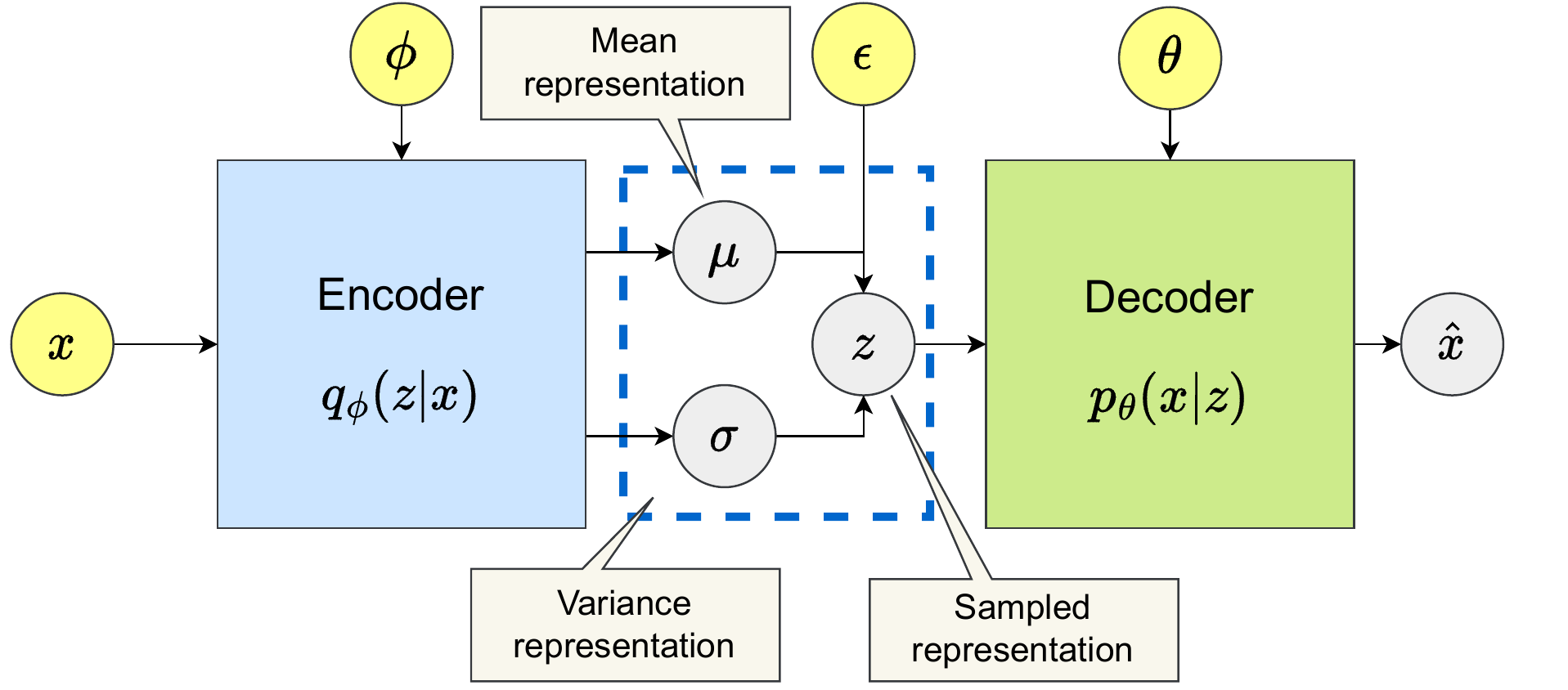}
    \caption{The structure of a VAE}
    \label{fig:vae-archi}
\end{figure}

This section presents a concise illustration of what mean, variance, and sampled representations are.
As shown in~\Figref{fig:vae-archi}, the mean, variance and sampled representations are the last 3 layers of the encoder,
where the sampled representation, $\rvz$, is the input of the decoder.

    \section{Passive variables and posterior collapse}\label{sec:app-collapse}
As discussed in~\Secref{subsec:bg-VAEs}, passive variables appear in latent representations
of VAEs in two cases: polarised regime and posterior collapse.
In well-behaved VAEs (i.e., in the case of polarised regime), passive variables arise when the number of latent dimensions is larger
than the number of latent variables needed by the VAE to encode latent representations. These
passive variables contribute to lowering the regularisation loss term of~\Eqref{eq:elbo} without increasing
the reconstruction loss.
However, passive variables can also be encountered as part of a pathological state where the reconstruction loss
is very high and the regularisation loss is pushed towards zero (i.e., when posterior collapse takes place).
This issue can happen for various reasons~\citep{Dai2020}, but is clearly distinct from
the polarised regime as the reconstruction loss is very high and the latent representations
contain little to no active variables.

In both cases, passive variables are very different in mean and sampled representations, due to the sampling process
$\rvz \sim \vmu + \epsilon \sqrt{\evsigma \mI}$, where $\epsilon \sim \N(0,\mI)$, $\vmu$ is the mean representation and $\evsigma \mI$ the diagonal matrix of
the variance representations.
For the regularisation term to be low, one needs to create passive variables, that is,
as many dimensions of $\rvz$ as close as possible to $\N(0, \mI)$.
This can easily be done by setting some elements of $\vmu$ to 0 and their corresponding variance to 1.
As a result, passive variables, when observed over multiple data examples will have a mean of
0 in the mean and sampled representations.
However, their variance will be close to 0 in the mean representations, and close to 1 in their sampled counterpart~\citep{Rolinek2019,Bonheme2021}.
    \section{Why not use variable type instead of IDE for FONDUE?}\label{sec:app-var-type}
As passive variables are easy to detect, one could wonder why they were not used directly to determine the number of
latent dimensions instead of comparing IDEs of models trained for a few epochs multiple times.
For example, it would be quicker to train one model with a large number of latent variables for a few epochs
and retrieve the number of active (or active and mixed) variables detected, as for example, illustrated in~\Algref{alg:fondue-var-type}.

\paragraph{How does~\Algref{alg:fondue-var-type} work?}
We define the initial number of latent variables as twice the data IDE.
Then, if we want to have enough dimensions for active and mixed variables, we double the number of latent variables until we find at least one passive variable
and return the sum of active and mixed variables as the chosen number of latent dimensions.
If we want instead to have only active variables, we double the number of latent variables until we find at least either one passive or mixed variable and return
the number of active variables as the chosen number of latent dimensions.

\paragraph{Why use~\Algref{alg:fondue} instead?}
As shown in~\Tableref{table:var-type}, the identification of variable types displays a high variance during early training,
which generally makes~\Algref{alg:fondue-var-type} less reliable than~\Algref{alg:fondue} for equivalent
computation time.
In addition to this instability, the numbers of latent dimensions predicted by~\Algref{alg:fondue-var-type} in~\Tableref{table:var-type}
are far from optimal compared to~\Tableref{table:fondue-ides}. There is a large overestimation in Symsol and an underestimation in Celeba.
These issues may be explained by the fact that~\Algref{alg:fondue} is based on mean and sampled representations,
while~\Algref{alg:fondue-var-type} solely relies on variance representations, decreasing the stability during early training.
Moreover, mixed and passive variables are not discriminated correctly in early epochs, possibly for the same reasons, preventing
any modulation of compression/reconstruction quality that could be achieved with~\Algref{alg:fondue}.

\begin{algorithm}
    \caption{FONDUE with variable types}\label{alg:fondue-var-type}
    \begin{algorithmic}
         \Procedure{FONDUE-var}{$data\_ide, epochs, keep\_mixed$}
             \State $l \gets 2 * data\_ide$
             \State $n \gets -1$

             \While{$n < 0$}
                \State $vae \gets train\_VAE(dim=l, n\_epochs=epochs)$
                \State $av, mv, pv \gets variable\_types(vae)$ \Comment{Number of active, mixed and passive variables}
                \If{$pv > 0$ and $keep\_mixed$}
                    \State $n \gets av + mv$
                \ElsIf{($mv > 0$ or $pv > 0$) and not $keep\_mixed$}
                    \State $n \gets av$
                \Else
                    \State $l \gets l * 2$
                \EndIf
             \EndWhile

             \State \textbf{return} $n$
         \EndProcedure
    \end{algorithmic}
\end{algorithm}

\begin{table}[ht!]
    \centering
    \caption{Number of latent variables obtained with FONDUE-var.
    The results are averaged over 5 seeds and computation times are reported for NVIDIA A100 GPUs.}
    \label{table:var-type}
    \begin{tabular}{ l l l l l}
        \hline
        Dataset & Dimensionality (avg $\pm$ SD) & Time/run & Models trained & Epochs/training\\
        \hline
        \rule{0pt}{2.6ex}Symsol & 14 $\pm$ 1.2 & 2 min & 3 & 1\\
        Symsol & 17 $\pm$ 1.7 & 3 min & 3 & 2\\
        Symsol & 15 $\pm$ 1.1 & 10 min & 3 & 5\\
        dSprites & 9.2 $\pm$ 1.3 & 4 min & 1 & 1\\
        dSprites & 8.8 $\pm$ 1.2 & 8 min & 1 & 2\\
        dSprites & 9.6 $\pm$ 0.5 & 20 min & 1 & 5\\
        Celeba & 38.6 $\pm$ 2.6 & 1 min & 1 & 1\\
        Celeba & 38.6 $\pm$ 0.4 & 2 min & 1 & 2\\
        Celeba & 41.2 $\pm$ 0.4 & 6 min & 1 & 5\\
        \hline
    \end{tabular}
\end{table}

\end{document}